\documentclass[10pt,twocolumn,letterpaper]{article}

\usepackage{times}
\usepackage{epsfig}
\usepackage{graphicx}
\usepackage{float}
\usepackage{wrapfig}
\usepackage{amsmath,amssymb,amsthm}
\usepackage{algorithm,algorithmicx,algpseudocode}
\usepackage{bm,xspace}
\usepackage{comment}
\usepackage{verbatim}
\usepackage{multirow}
\usepackage{balance}
\usepackage{url}
\usepackage{booktabs}
\usepackage{etoolbox,siunitx}
\usepackage{calc}
\usepackage{pifont,hologo}
\usepackage[usenames, dvipsnames]{color}
\usepackage{nicefrac}

\newcommand{\ra}[1]{\renewcommand{\arraystretch}{#1}}

\newcommand*\samethanks[1][\value{footnote}]{\footnotemark[#1]}

\usepackage[super]{nth}
\usepackage{nicefrac}
\robustify\bfseries

\def\gg{\mathbf{g}}

\def\pp{\mathbf{p}}
\def\qq{\mathbf{q}}

\def\uu{\mathbf{u}}

\def\FF{\mathbf{F}}

\def\HH{\mathbf{H}}
\def\II{\mathbf{I}}

\def\MM{\mathbf{M}}

\def\pP{\mathcal{P}}

\def\vV{\mathcal{V}}

\def\trans{^{\top}}

\newcommand{\ColorMapCircle}{\ding{108}}

\DeclareMathSymbol{@}{\mathord}{letters}{"3B}

\newcommand\timess{\mathbin{\!\times\!}}

\newcommand\mypara[1]{\vspace{1mm}\noindent\textbf{#1}}

\def\latex/{\LaTeX}
\def\bibtex/{\hologo{BibTeX}}

\usepackage{cvpr}
\usepackage{times}
\usepackage{epsfig}
\usepackage{graphicx}
\usepackage{amsmath}
\usepackage{amssymb}
\makeatletter
\@namedef{ver@everyshi.sty}{}
\makeatother
\usepackage{tikz}
\usepackage{pgfplots}
\usepackage[abs]{overpic}

\usepackage[pagebackref=true,breaklinks=true,letterpaper=true,colorlinks,bookmarks=false]{hyperref}

\cvprfinalcopy 

\def\cvprPaperID{144} 

\begin{document}

\title{Tangent Convolutions for Dense Prediction in 3D}

\author{Maxim Tatarchenko\thanks{Equal contribution.}\\
University of Freiburg\\
\and
Jaesik Park\samethanks{}\\
Intel Labs\\
\and
Vladlen Koltun\\
Intel Labs\\
\and
Qian-Yi Zhou\\
Intel Labs\\
}

\maketitle

\begin{abstract}
We present an approach to semantic scene analysis using deep convolutional networks. Our approach is based on tangent convolutions -- a new construction for convolutional networks on 3D data. In contrast to volumetric approaches, our method operates directly on surface geometry. Crucially, the construction is applicable to unstructured point clouds and other noisy real-world data. We show that tangent convolutions can be evaluated efficiently on large-scale point clouds with millions of points. Using tangent convolutions, we design a deep fully-convolutional network for semantic segmentation of 3D point clouds, and apply it to challenging real-world datasets of indoor and outdoor 3D environments. Experimental results show that the presented approach outperforms other recent deep network constructions in detailed analysis of large 3D scenes.
\end{abstract}


\section{Introduction}
\label{sec:introduction}

Methods that utilize convolutional networks on 2D images dominate modern computer vision.
A key contributing factor to their success is efficient local processing based on the convolution operation.
2D convolution is defined on a regular grid, a domain that supports extremely efficient implementation.
This in turn enables using powerful deep architectures for processing large datasets at high resolution.

When it comes to analysis of large-scale 3D scenes, a straightforward extension of this idea is volumetric convolution on a voxel grid~\cite{maturana15,wu15,dai17}. However, voxel-based methods have limitations, including a cubic growth rate of memory consumption and computation time.
For this reason, voxel-based ConvNets operate on low-resolution voxel grids that limit their prediction accuracy.
The problem can be alleviated by octree-based techniques that define a ConvNet on an octree and enable processing somewhat higher-resolution volumes (e.g., up to $256^3$ voxels)~\cite{riegler17,wang17,Hane2017,Riegler2017:3DV,Tatarchenko2017}.
Yet even this may be insufficient for detailed analysis of large-scale scenes.

On a deeper level, both efficient and inefficient voxel-based methods treat 3D data as \emph{volumetric} by exploiting 3D convolutions that integrate over volumes. In reality, data captured by 3D sensors such as RGB-D cameras and LiDAR typically represent \emph{surfaces}: 2D structures embedded in 3D space. (This is in contrast to truly volumetric 3D data, as encountered for example in medical imaging.) Classic features that are used for the analysis of such data are defined in terms that acknowledge the latent surface structure, and do not treat the data as a volume~\cite{JohnsonHebert1999,Frome2004,Rusu2009}.

\begin{figure}
\centering
\begin{tabular}{@{}c@{}}
\includegraphics[width=1\linewidth, trim={0 0 0 40mm}, clip]{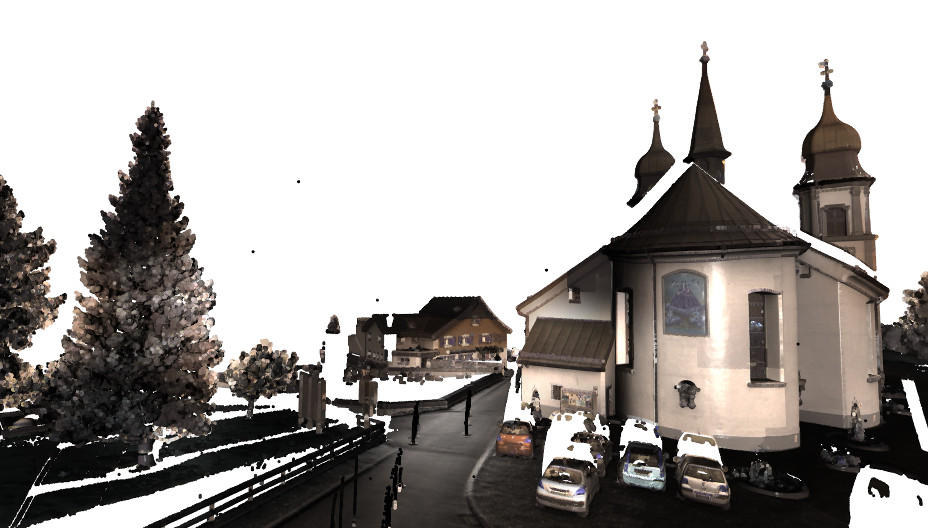} \\
\includegraphics[width=1\linewidth, trim={0 0 0 40mm}, clip]{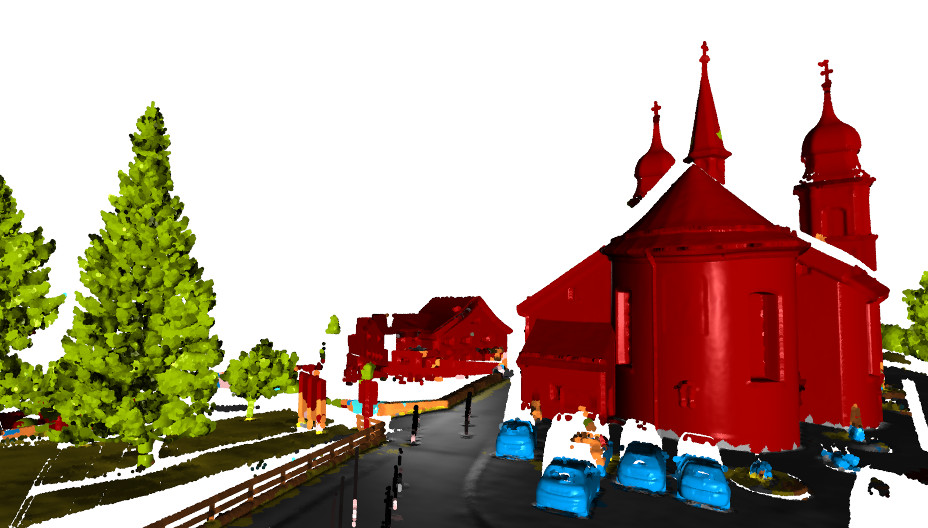}
\end{tabular}
\caption{Convolutional networks based on tangent convolutions can be applied to semantic analysis of large-scale scenes, such as urban environments. Top: point cloud from the Semantic3D dataset. Bottom: semantic segmentation produced by the presented approach.}
\label{fig:teaser}
\end{figure}

The drawbacks of voxel-based methods are known in the research community.
A number of recent works argue that volumetric data structures are not the natural substrate for 3D ConvNets, and propose alternative designs based on unordered point sets~\cite{qi17}, graphs~\cite{simonovsky17}, and sphere-type surfaces~\cite{maron17}. Unfortunately, these methods come with their own drawbacks, such as limited sensitivity to local structure or restrictive topological assumptions.

We develop an alternative construction for convolutional networks on surfaces, based on the notion of \emph{tangent convolution}. This construction assumes that the data is sampled from locally Euclidean surfaces. The latent surfaces need not be known, and the data can be in any form that supports approximate normal vector estimation, including point clouds, meshes, and even polygon soup. (The same assumption concerning normal vector estimation is made by both classic and contemporary geometric feature descriptors~\cite{JohnsonHebert1999,Frome2004,Rusu2009,Tombari2010b,Salti2014,khoury17}.)
The tangent convolution is based on projecting local surface geometry on a tangent plane around every point. This yields a set of tangent images. Every tangent image is treated as a regular 2D grid that supports planar convolution. The content of all tangent images can be precomputed from the surface geometry, which enables efficient implementation that scales to large datasets, such as urban environments.

Using tangent convolution as the main building block, we design a U-type network for dense semantic segmentation of point clouds.
Our proposed architecture is general and can be applied to analysis of large-scale scenes.
We demonstrate its performance on three diverse real-world datasets containing indoor and outdoor environments.
A semantic segmentation produced by a tangent convolutional network is shown in Figure~\ref{fig:teaser}.

\section{Related Work}
\label{sec:related_work}
Dense prediction in 3D, including semantic point cloud segmentation, has a long history in computer vision.
Pioneering methods work on aerial LiDAR data and are based on hand-crafted features with complex classifiers on top~\cite{charaniya04,chehata09,golovinskiy09}.
Such approaches can also be combined with high-level architectural rules~\cite{martinovic15}.
A popular line of work exploits graphical models, including conditional random fields~\cite{munoz09,floros12,anand13,wu14,hermans14,kundu14,vineet15}.
Related formulations have also been proposed for interactive 3D segmentation~\cite{Valentin2015,miksik15}.

More recently, the deep learning revolution in computer vision has spread to consume 3D data analysis. A variety of methods that tackle 3D data using deep learning techniques have been proposed. They can be considered in terms of the underlying data representation.

A common representation of 3D data for deep learning is a voxel grid. Deep networks that operate on voxelized data have been applied to shape classification~\cite{maturana15,wu15,qi16}, semantic segmentation of indoor scenes~\cite{dai17}, and biomedical recordings~\cite{cicek16,chen16}. Due to the cubic complexity of voxel grids, these methods can only operate at low resolution~-- typically not more than $64^3$~-- and have limited accuracy. Attempting to overcome this limitation, researchers have proposed representations based on hierarchical spatial data structures such as octrees and kd-trees~\cite{riegler17,wang17,Hane2017,Riegler2017:3DV,Gadelha2017:BMVC,KlokovLempitsky2017,Tatarchenko2017}, which are more memory- and computation-efficient, and can therefore handle higher resolutions. An alternative way of increasing the accuracy of voxel-based techniques is to add differentiable post-processing, modeled upon the dense CRF~\cite{KrahenbuhlKoltun2011,tchapmi17}.

Other applications of deep networks consider RGB-D images, which can be treated with fully-convolutional networks~\cite{Gupta2015,li16:ECCV,mccormac17} and graph neural networks~\cite{qi17iccv}. These approaches support the use of powerful pretrained 2D networks, but are not generally applicable to unstructured point clouds with unknown sensor poses. Attempting to address this issue, Boulch et al.~\cite{boulch17} train a ConvNet on images rendered from point clouds using randomly placed virtual cameras. In a more controlled setting with fixed camera poses, multi-view methods are successfully used for shape segmentation~\cite{kalogerakis17}, shape recognition~\cite{su15,qi16}, and shape synthesis~\cite{Gadelha2017:3DV,Lun2017,Kar2017,tatarchenko16}. Our approach can be viewed as an extreme multi-view approach in which a virtual camera is associated with each point in the point cloud. A critical problem that we address is the efficient and scalable implementation of this approach, which enables its application to dense point clouds of large-scale indoor and outdoor environments.

Qi et al.~\cite{qi17} propose a network for analysing unordered point sets, which is based on independent point processing combined with global context aggregation through max-pooling. Since the communication between the points is quite weak, this approach experiences difficulties when applied to large-scale scenes with complex layouts.


There is a variety of more exotic deep learning formulations for 3D analysis that do not address large-scale semantic segmentation of whole scenes but provide interesting ideas.
Yi et al.~\cite{yi17} consider shape segmentation in the spectral domain by synchronizing eigenvectors across models.
Masci et al.~\cite{masci15} and Boscaini et al.~\cite{Boscaini2016} design ConvNets for Riemannian manifolds and use them to learn shape correspondences.
Sinha et al.~\cite{sinha16} perform shape analysis on geometry images.
Simonovsky et al.~\cite{simonovsky17} extend the convolution operator from regular grids to arbitrary graphs and use it to design shape classification networks.
Li et al.~\cite{li16:NIPS} introduce Field Probing Neural Networks which respect the underlying sparsity of 3D data and are used for efficient feature extraction.
Tulsiani et al.~\cite{tulsiani17} approximate 3D models with volumetric primitives in an end-to-end differentiable framework, and use this representation for solving several tasks.
Maron et al.~\cite{maron17} design ConvNets on surfaces for sphere-type shapes.

Overall, most existing 3D deep learning systems either rely on representations that do not support general scene analysis, or have poor scalability. As we will show, deep networks based on tangent convolutions scale to millions of points and are suitable for detailed analysis of large scenes.

\section{Tangent Convolution}
\label{sec:parametrization}

In this section we formally introduce tangent convolutions. All derivations are provided for point clouds, but they can easily be applied to any type of 3D data that supports surface normal estimation, such as meshes.

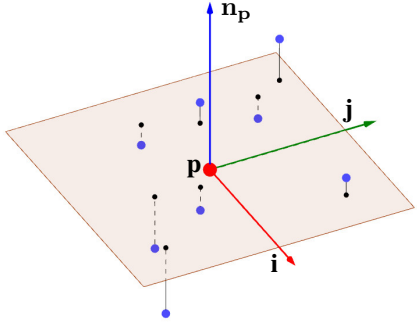
\begin{figure}
\centering
\resizebox{0.34\textwidth}{!}{
\begin{overpic}[width=0.7\linewidth]{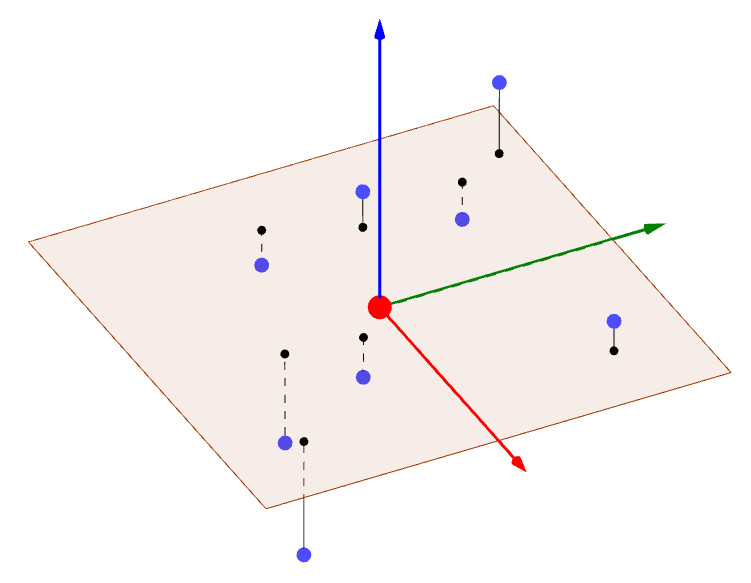}
 \put (88,120) {\normalsize $\mathbf{n}_\mathbf{p}$}
 \put (107,22) {\normalsize $\textbf{i}$}
 \put (135,81) {\normalsize $\textbf{j}$}
 \put (75,60) {\normalsize $\textbf{p}$}
\end{overpic}
}
\caption{Points $\mathbf{q}$ (blue) from the local neighborhood of a point $\mathbf{p}$ (red) are projected onto the tangent image.}
\label{fig:tangent_image}
\end{figure}

\mypara{Convolution with a continuous kernel.}
Let $\pP = \{\pp\}$ be a point cloud, and let $F(\pp)$ be a discrete scalar function that represents a signal defined over $\pP$.
$F(\pp)$ can encode color, geometry, or abstract features from intermediate network layers.
In order to convolve $F$, we need to extend it to a continuous function.
Conceptually, we introduce a virtual orthogonal camera for $\mathbf{p}$.
It is configured to observe $\mathbf{p}$ along the normal $\mathbf{n}_\mathbf{p}$.
The image plane of this virtual camera is the tangent plane $\pi_\mathbf{p}$ of $\mathbf{p}$.
It parameterizes a virtual image that can be represented as a continuous signal $S(\mathbf{u})$, where $\mathbf{u} \in \mathbb{R}^2$ is a point in $\pi_\mathbf{p}$. We call $S$ a tangent image.

The tangent convolution at $\pp$ is defined as
\begin{equation}
X(\pp) = \int_{\pi_\mathbf{p}} c(\mathbf{u}) S(\mathbf{u})\: d\mathbf{u},
\label{eq:convolution_basic}
\end{equation}
where $c(\mathbf{u})$ is the convolution kernel. We now describe how $S$ is computed from $F$.

\mypara{Tangent plane estimation.}
For each point $\mathbf{p}$ we estimate the orientation of its camera image using local covariance analysis.
This is a standard procedure~\cite{Salti2014} but we summarize it here for completeness.
Consider a set of points ${\mathbf{q}}$ from a spherical neighborhood of $\pp$, such that $\|\mathbf{p}-\mathbf{q}\| < R$.
The orientation of the tangent plane is determined by the eigenvectors of the covariance matrix $\mathbf{C} = \sum_\mathbf{q} \mathbf{r}\mathbf{r}\trans$, where $\mathbf{r} = \mathbf{q} - \mathbf{p}$.
The eigenvector of the smallest eigenvalue defines the estimated surface normal $\mathbf{n}_\mathbf{p}$, and the other two eigenvectors $\mathbf{i}$ and $\mathbf{j}$ define the 2D image axes that parameterize the tangent image.

\begin{figure}
\centering
\begin{tabular}{@{}c@{\hspace{1mm}}c@{\hspace{1mm}}c@{\hspace{1mm}}c@{}}
\includegraphics[width=0.24\linewidth]{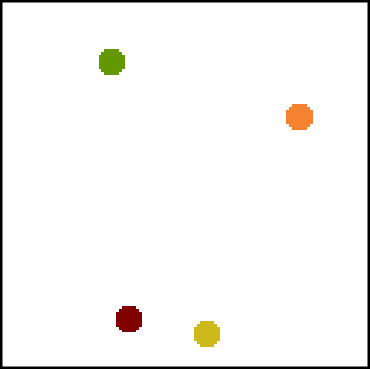} &
\includegraphics[width=0.24\linewidth]{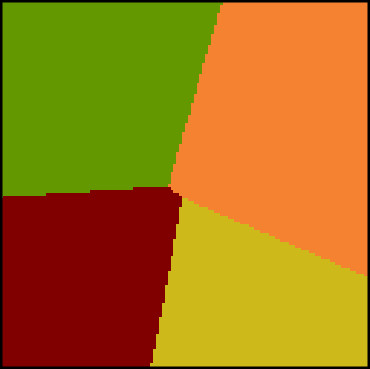} &
\includegraphics[width=0.24\linewidth]{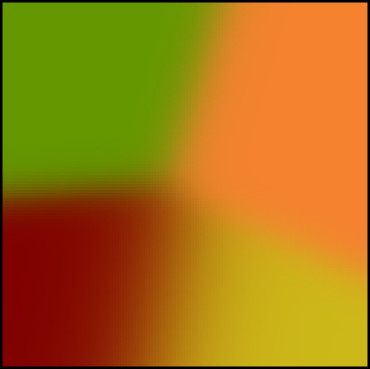} &
\includegraphics[width=0.24\linewidth]{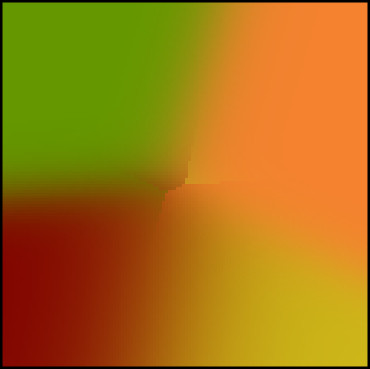}\\
\small (a) & \small (b) & \small (c) & \small (d)
\end{tabular}
\vspace{1mm}
\caption{Signals from projected points (a) can be interpolated using one of the following schemes: nearest neighbor (b), full Gaussian mixture (c), and Gaussian mixture with top-3 neighbors (d).}
\label{fig:signal_interpolation}
\end{figure}

\mypara{Signal interpolation.}
Now our goal is to estimate image signals $S(\mathbf{u})$ from point signals $F(\mathbf{q})$.
We begin by projecting the neighbors $\qq$ of $\pp$ onto the tangent image, which yields a set of projected points $\mathbf{v}=(\mathbf{r}\trans \mathbf{i}, \mathbf{r}\trans \mathbf{j})$.
This is illustrated in Figure~\ref{fig:tangent_image}.
We define
\begin{equation}
S(\mathbf{v}) = F(\mathbf{q}).
\label{eq:feature_assign}
\end{equation}
As shown in Figure~\ref{fig:tangent_image} and Figure~\ref{fig:signal_interpolation}(a), points $\mathbf{v}$ are scattered on the image plane. We thus need to interpolate their signals in order to estimate the full function $S(\mathbf{u})$ over the tangent image:
\begin{equation}
S(\mathbf{u}) = \sum_{\mathbf{v}}\big(w(\mathbf{u},\mathbf{v}) \cdot S(\mathbf{v})\big),
\label{eq:linear_approximation}
\end{equation}
where $w(\mathbf{u}, \mathbf{v})$ is a kernel weight that satisfies ${\sum_{\mathbf{v}}w = 1}$.
We consider two schemes for signal interpolation: nearest neighbor and Gaussian kernel mixture. These schemes are illustrated in Figure~\ref{fig:signal_interpolation}.
In the nearest neighbor (NN) case,
\begin{equation}
w(\mathbf{u}, \mathbf{v}) =
	\begin{cases}
    1 & \text{if } \mathbf{v} \text{ is } \mathbf{u} \text{'s NN},\\
	0 & \text{otherwise}.
	\end{cases}
\label{eq:nn}
\end{equation}
In the Gaussian kernel mixture case,
\begin{equation}
w(\mathbf{u}, \mathbf{v}) = \frac{1}{A}\exp\left({-\frac{\|\mathbf{u}-\mathbf{v}\|^2}{\sigma^2}}\right),
\label{eq:gaussian}
\end{equation}
where $A$ normalizes the weights such that ${\sum_{\mathbf{v}}w = 1}$. More sophisticated signal interpolation schemes can be considered, but we have not observed a significant effect of the interpolation scheme on empirical performance and will mostly use simple nearest-neighbor estimation.

Finally, if we rewrite Equation~\eqref{eq:convolution_basic} using the definitions from Equations~\eqref{eq:feature_assign}~and~\eqref{eq:linear_approximation}, we get the formula for the tangent convolution:
\begin{equation}
X(\pp) = \int_{\pi_{\pp}} c(\mathbf{u}) \cdot \sum_{\mathbf{v}}\big(w(\mathbf{u}, \mathbf{v}) \cdot F(\mathbf{q})\big) \: d\mathbf{u}.
\label{eq:convolution}
\end{equation}

Note that the role of the tangent image is increasingly implicit: it provides the domain for $\uu$ and figures in the evaluation of the weights $w$, but otherwise it need not be explicitly maintained. We will build on this observation in the next section to show that tangent convolutions can be evaluated efficiently at scale, and can support the construction of deep networks on point clouds with millions of points.

\section{Efficiency}
\label{sec:efficiency}
In this section we describe how the tangent convolution defined in Section~\ref{sec:parametrization} can be computed efficiently.
In practice, the tangent image is treated as a discrete function on a regular $l \timess l$ grid.
Elements $\mathbf{u}$ are pixels in this virtual image.
The convolution kernel $c$ is a discrete kernel applied onto this image.
Let us first consider the nearest-neighbor signal interpolation scheme introduced in Equation~\eqref{eq:nn}.
We can rewrite Equation~\eqref{eq:convolution} as
\begin{equation}
X(\pp) = \sum_\mathbf{u}\Big(c(\mathbf{u}) \cdot F\big(g(\mathbf{u})\big)\Big),
\end{equation}
where $g(\mathbf{u})$ is a selection function that returns a point which projects to the nearest neighbor of $\mathbf{u}$ on the image plane.
Note that $g$ only depends on the point cloud geometry and does not depend on the signal $F$. This allows us to precompute $g$ for all points.

From here on, we employ standard ConvNet terminology and proceed to show how to implement a convolutional layer using tangent convolutions. Our goal is to convolve an input feature map $\FF_{in}$ of size $N \timess C_{in}$ with a set of weights $W$ to produce an output feature map $\FF_{out}$ of size $N \timess C_{out}$, where $N$ is the number of points in the point cloud, while $C_{in}$ and $C_{out}$ denote the number of input and output channels respectively.
For implementation, we unroll 2D tangent images and convolutional filters of size $l \timess l$ into 1D vectors of size $1 \timess L$, where $L = l^2$. From then on, we compute 1D convolutions. Note that such representation of a 2D tangent convolution as a 1D convolution is not an approximation: the results of the two operations are identical.

\begin{figure}
\centering
\begin{overpic}[width=\linewidth]{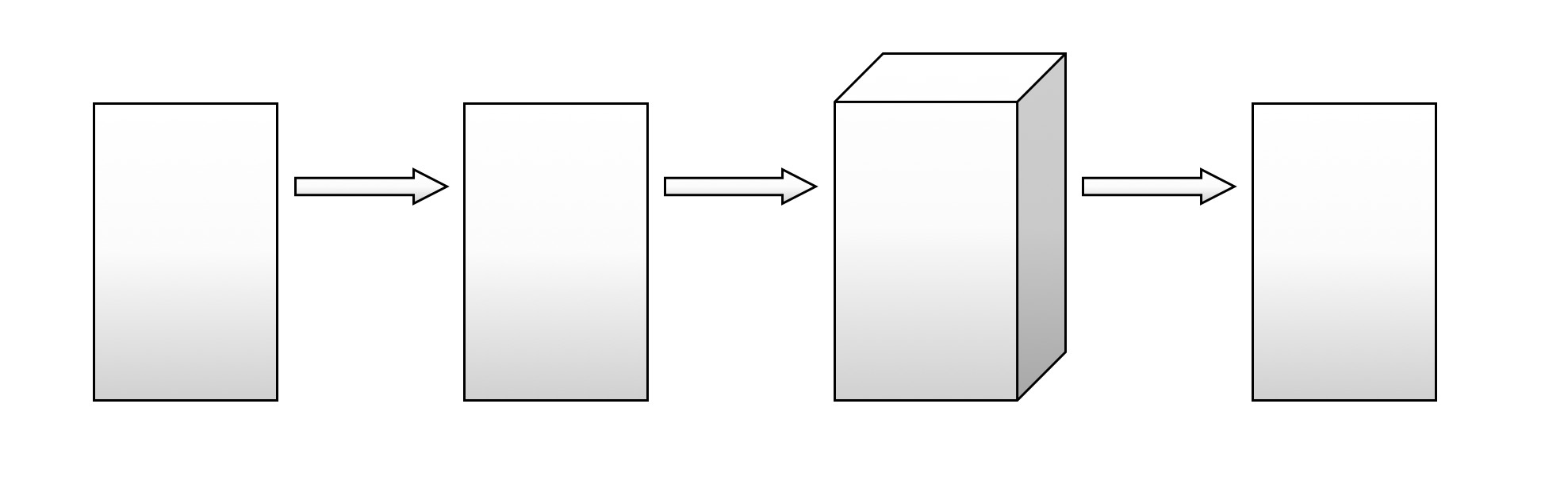}
 \put (6,30) {\scriptsize $N$}
 \put (62,30) {\scriptsize $N$}
 \put (118,30) {\scriptsize $N$}
 \put (22,62) {\scriptsize $C_{in}$}
 \put (117,65) {\scriptsize $C_{in}$}
 \put (196,62) {\scriptsize $C_{out}$}
 \put (83,61.5) {\scriptsize $L$}
 \put (139,61.5) {\scriptsize $L$}
 \put (20,35) {\large $\mathbf{F_{in}}$}
 \put (82.5,35) {\large $\mathbf{I}$}
 \put (135,35) {\large $\mathbf{M}$}
 \put (192,35) {\large $\mathbf{F_{out}}$}
 \put (48,53) {\footnotesize $g(\mathbf{u})$}
 \put (98,53) {\scriptsize $F(g(\mathbf{u}))$}
 \put (168,53) {\scriptsize conv}
\end{overpic}
\caption{Efficient evaluation of a convolutional layer built on tangent convolutions.}
\label{fig:implementation}
\end{figure}
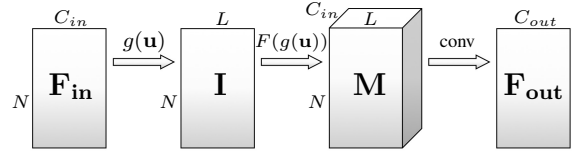

We start by precomputing the function $g$, which is represented as an $N \timess L$ index matrix $\II$.
Elements of $\II$ are indices of the corresponding tangent-plane nearest-neighbors in the point cloud.
Using $\II$, we gather input signals (features) into an intermediate tensor $\MM$ of size $N \timess L \timess C_{in}$.
This tensor is convolved with a flattened set of kernels $W$ of size $1 \timess L$, which yields the output feature map $\FF_{out}$. This process is illustrated in Figure~\ref{fig:implementation}.

Consider now the case of signal interpolation using Gaussian kernel mixtures.
For efficiency, we only consider the set of top-$k$ neighbors for each point, denoted $NN_k$.
An example image produced using the Gaussian kernel mixture scheme with top-3 neighbors is shown in Figure~\ref{fig:signal_interpolation}(d).
Equation~\eqref{eq:gaussian} turns into
\begin{equation}
w(\mathbf{u}, \mathbf{v}) =
	\begin{cases}
    \frac{1}{A}\exp\left({-\frac{\|\mathbf{u}-\mathbf{v}\|^2}{\sigma^2}}\right) & \text{if $\mathbf{v}$} \in NN_k\\
	0 & \text{otherwise},
	\end{cases}
\label{eq:gaussian2}
\end{equation}
where $A$ normalizes weights such that $\sum_\mathbf{v}w=1$. With this approximation, each pixel $\mathbf{u}$ has at most $k$ non-zero weights, denoted by $w_{1..k}(\mathbf{u})$. Their corresponding selection functions are denoted by $g_{1..k}(\mathbf{u})$. Both the weights and the selection functions are independent of the signal $F$, and are thus precomputed. Equation~\eqref{eq:convolution} becomes
\begin{eqnarray}
	X(\pp) &=& \sum_\mathbf{u}\Big(c(\mathbf{u}) \cdot \sum_{i=1}^{k}\big(w_{i}(\mathbf{u}) \cdot F(g_i(\mathbf{u}))\big)\Big)\\
	&=&\sum_{i=1}^{k}\sum_\mathbf{u}\Big(w_{i}(\mathbf{u})\cdot c(\mathbf{u}) \cdot F(g_i(\mathbf{u}))\Big).\label{eq:convolution_gaussian}
\end{eqnarray}


As with the nearest-neighbor signal interpolation scheme, we represent the precomputed selection functions $g_i$ as $k$ index matrices $\II_{i}$ of size $N \timess L$. These index matrices are used to assemble $k$ intermediate signal tensors $\MM_i$ of size $N \timess L \timess C_{in}$. Additionally, we collate the precomputed weights into $k$ weight matrices $\HH_i$ of size $N \timess L$. They are used to compute the weighted sum ${\MM=\sum_i \HH_i \odot \MM_i}$, which is finally convolved with the kernel $W$.

We implemented\footnote{\url{https://github.com/tatarchm/tangent_conv}} the presented construction in TensorFlow~\cite{abadi16}.
It consists entirely of differentiable atomic operations, thus backpropagation is done seamlessly using the automatic differentiation functionality of the framework.

\section{Additional Ingredients}
\label{sec:ingredients}
In this section we introduce additional ingredients that are required to construct a convolutional network for point cloud analysis.





\subsection{Multi-scale analysis}

\mypara{Pooling.}
Convolutional networks commonly use pooling to aggregate signals over larger spatial regions.
We implement pooling in our framework via hashing onto a regular 3D grid.
Points that are hashed onto the same grid point pool their signals.
The spacing of the grid determines the pooling resolution.
Consider points $\pP = \{\pp\}$ and corresponding signal values $\{F(\pp)\}$.
Let $\gg$ be a grid point and let $\vV_\gg$ be the set of points in $\pP$ that hash to $\gg$. (The hash function can be assumed to be simple quantization onto the grid in each dimension.) Assume that $\vV_\gg$ is not empty and consider average pooling. All points in $\vV_\gg$ and their signals are pooled onto a single point:
\begin{equation}
\resizebox{0.9\width}{!}{$\displaystyle{
\pp'_\gg = \frac{1}{|\vV_\gg|}\sum_{\pp \in \vV_\gg}\pp
}$}
\quad\ \text{and} \quad\
\resizebox{0.9\width}{!}{$\displaystyle{
F'(\pp'_{\gg}) = \frac{1}{|\vV_\gg|}\sum_{\pp \in \vV_\gg}F(\pp).
}$}
\label{eq:2}
\end{equation}
In a convolutional network based on tangent convolutions, we pool using progressively coarser grids. Starting with some initial grid resolution (5cm in each dimension, say), each successive pooling layer increases the step of the grid by a factor of two (to 10cm, then 20cm, etc.).
Such hashing also alleviates the problem of non-uniform point density. As a result, we can select the neighborhood radius for the convolution operation globally for the entire dataset.

After each pooling layer, the radius $r$ that is used to estimate the tangent plane and the pixel size of the virtual tangent image are doubled accordingly. Thus the resolution of all tangent images decreases in step with the resolution of the point cloud. Note that the downsampled point clouds produced by pooling layers are independent of the signals defined over them. The downsampled point clouds, the associated tangent planes, and the corresponding index and weight functions can thus all be precomputed for all layers in the convolutional network: they need only be computed once per pooling layer.

The implementation of a pooling layer is similar in spirit to that of a convolutional layer described in Section~\ref{sec:efficiency}.
Consider an input feature map $\FF_{in}$ of size $N_{in} \timess C$.
Using grid hashing, we assemble an index matrix $\II$ of size $N_{out} \timess 8$, which contains indices of points that hash to the same grid point. Assuming that we decrease the grid resolution by a factor of 2 in each dimension in each pooling layer, the number of points that hash to the same grid point will be at most 8 in general. (For initialization, we quantize the points to some base resolution.)
Using $\II$, we assemble an intermediate tensor of size $N_{out} \timess 8 \timess C$.
We pool this tensor along the second dimension according to the pooling operator (max, average, etc.), and thus obtain an output feature map $\FF_{out}$ of size $N_{out} \timess C$.

Note that all stages in this process have linear complexity in the number of points. Although points are hashed onto regular grids, the grids themselves are never constructed or represented. Hashing is performed via modular arithmetic on individual point coordinates, and all data structures have linear complexity in the number of points, independent of the extent of the point set or the resolution of the grid.

\mypara{Unpooling.}
The unpooling operation has an opposite effect to pooling: it distributes signals from points in a low-resolution feature map $\FF_{in}$ onto points in a higher-resolution feature map $\FF_{out}$. Unpooling reuses the index matrix from the corresponding pooling operation. We copy features from a single point in a low-resolution point cloud to multiple points from which the information was aggregated during pooling.

\subsection{Local distance feature}

So far, we have considered signals that could be expressed in terms of a scalar function $F(\qq)$ with a well-defined value for each point $\qq$.
This holds for color, intensity, and abstract ConvNet features.
There is, however, a signal that cannot be expressed in such terms and needs special treatment. This signal is distance to the tangent plane $\pi_{\pp}$. This local signal is calculated by taking the distance from each neighbor $\mathbf{q}$ to the tangent plane of $\pp$: ${d = (\mathbf{q}-\mathbf{p})\trans\mathbf{n}_{\pp}}$.

This signal is defined in relation to the point $\mathbf{p}$, therefore it cannot be directly plugged into the pipeline shown in Figure~\ref{fig:implementation}.
Instead, we precompute the distance images for every point.
Scattered signal interpolation is done in the same way as for scalar signals (Equation~\eqref{eq:linear_approximation}).
After assembling the intermediate tensor $\MM$ for the first convolutional layer, we simply concatenate these distance images as an additional channel in $\MM$.
The first convolutional layer generates a set of abstract features $\FF_{out}$ that can be treated as scalar signals from here on.

All precomputations are implemented using Open3D~\cite{zhou18}.

\section{Architecture}
\label{sec:architecture}
\begin{figure}
\centering
\resizebox{0.40\textwidth}{!}{
\begin{overpic}[width=\linewidth]{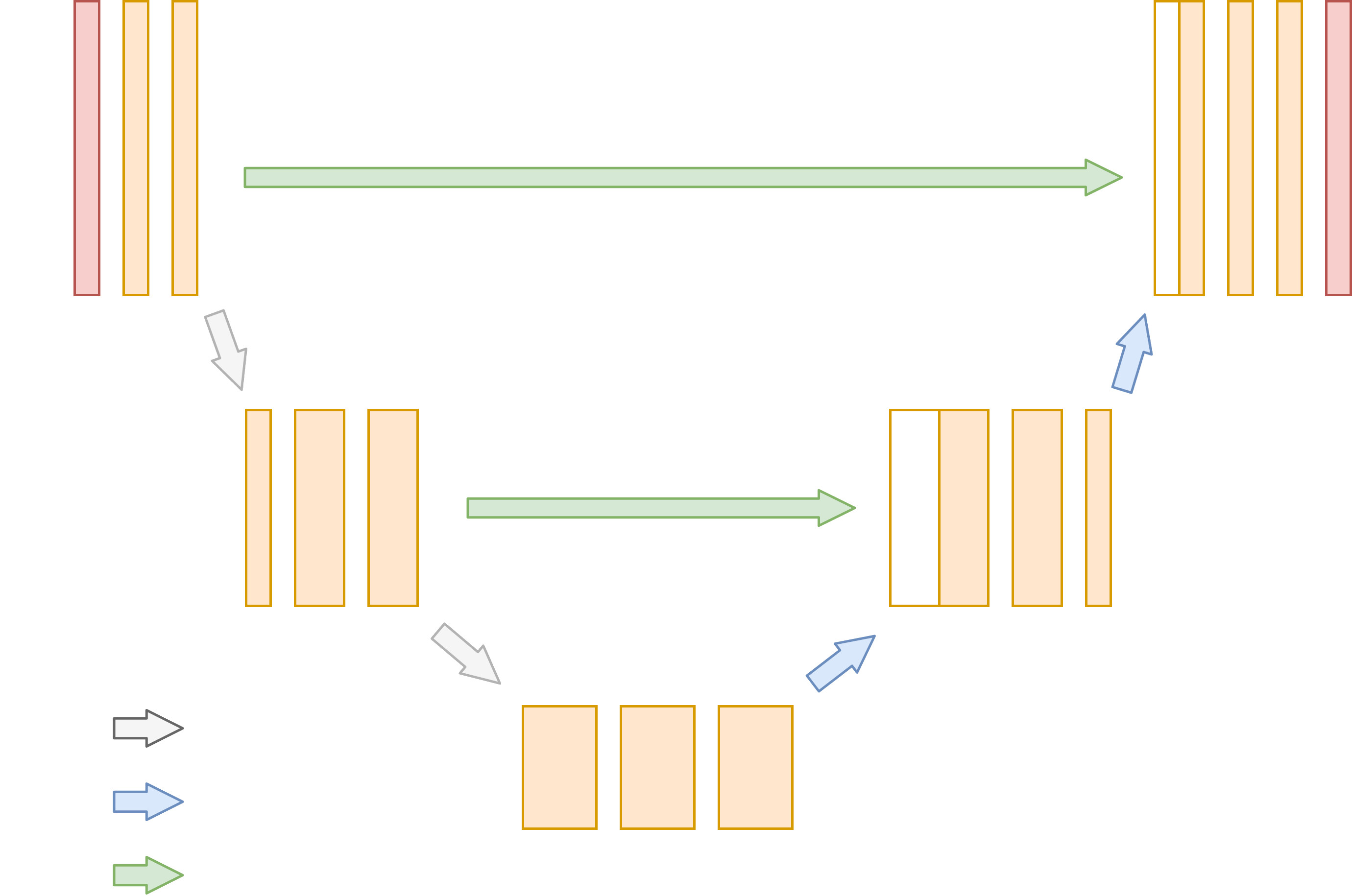}
 \put (12,101) {\tiny $m$}
 \put (20.5,101) {\tiny $32$}
 \put (29,101) {\tiny $32$}
 \put (42,46) {\tiny $32$}
 \put (53,46) {\tiny $64$}
 \put (66,46) {\tiny $64$}
 \put (95, 7) {\tiny $64$}
 \put (110, 7) {\tiny $128$}
 \put (129.5, 7) {\tiny $64$}
 \put (160,46) {\tiny $128$}
 \put (179,46) {\tiny $64$}
 \put (189.5,46) {\tiny $32$}
 \put (214,101) {\tiny $32$}
 \put (203.5,101) {\tiny $32$}
 \put (223,101) {\tiny $64$}
 \put (232.5,101) {\tiny $n$}
 \put (35,28) {\footnotesize pool}
 \put (35,14.5) {\footnotesize unpool}
 \put (35,1.5) {\footnotesize skip}
\end{overpic}
}
\vspace{1mm}
\caption{We use a fully-convolutional U-shaped network with skip connections. The network receives $m$-dimensional features as input and produces prediction scores for $n$ classes.}
\label{fig:architecture}
\end{figure}
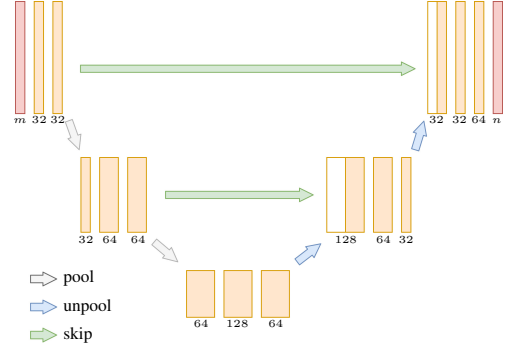

Using the ingredients introduced in the previous sections, we design an encoder-decoder network inspired by the U-net~\cite{Ronneberger2015}. The network architecture is illustrated in Figure~\ref{fig:architecture}.
It is a fully-convolutional network over a point cloud, where the convolutions are tangent convolutions. The encoder contains two pooling layers. The decoder contains two corresponding unpooling layers.
Encoder features are propagated to corresponding decoder blocks via skip-connections.
All layers except the last one use $3 \timess 3$ filters and are followed by Leaky ReLU with negative slope 0.2~\cite{Maas2013}.
The last layer uses $1 \timess 1$ convolutions to produce final class predictions.
The network is trained by optimizing the cross-entropy objective using the Adam optimizer with initial learning rate $10^{-4}$~\cite{KingmaBa2015}.

\mypara{Receptive field.}
The receptive field size of one convolutional layer is determined by the pixel size $r$ of the tangent image and the radius $R$ that is used to collect the neighbors of each point $\pp$. We set $R=2 r$, therefore the receptive field size of one layer is $R$.
After each pooling layer, $r$ is doubled.
The receptive field of an element in the network can be calculated by tracing the receptive fields of preceding layers.
With initial $r=5$cm, the receptive field size of elements in the final layer of the presented architecture is $4\cdot10 + 4\cdot20 + 2\cdot40=200\text{cm}$.

\section{Experiments}
\label{sec:results}
We evaluate the performance of the presented approach on the task of semantic 3D scene segmentation. Our approach is compared to recent deep networks for 3D data on three different datasets.

\subsection{Datasets and measures}

We conduct experiments on three large-scale datasets that contain real-world 3D scans of indoor and outdoor environments.

\mypara{Semantic3D}~\cite{hackel17} is a dataset of scanned outdoor scenes with over 3 billion points.
It contains 15 training and 15 test scenes annotated with 8 class labels.
Being unable to evaluate the baseline results on the official test server, we use our own train/test split:
Bildstein 1-3-5 are used for testing, the rest for training.

\mypara{Stanford Large-Scale 3D Indoor Spaces Dataset} (S3DIS)~\cite{armeni16} contains 6 large-scale indoor areas from 3 different buildings, with 13 object classes.
We use Area 5 for testing and the rest for training.

\mypara{ScanNet}~\cite{dai17} is a dataset with more than 1,500 scans of indoor scenes with 20 object classes collected using an \mbox{RGB-D} capture system.
We follow the standard train/test split provided by the authors.

\mypara{Measures.}
We report three measures: mean accuracy over classes (mA), mean intersection over union (mIoU), and overall accuracy (oA).
We build a full confusion matrix based on the entire test set, and derive the final scores from it.
Measures are evaluated over the original point clouds.
For approaches that produce labels over downsampled or voxelized representations, we map these predictions to the original point clouds via nearest-neighbor assignment.

Although we report oA for completeness, it is not a good measure for semantic segmentation. If there are dominant classes in the data (e.g., walls, floor, and ceiling in indoor scenes), making correct predictions for these but poor predictions over the other classes will yield misleadingly high oA scores.

\subsection{Baselines}

We compare our approach to three recent deep learning methods that operate on different underlying representations.
We have chosen reasonably general methods that have the potential to be applied to general scene analysis and have open-source implementations.
Our baselines are PointNet~\cite{qi17}, which operates on points, ScanNet~\cite{dai17}, which operates on low-resolution voxel grids, and OctNet~\cite{riegler17}, which operates on higher-resolution octrees.
We used the source code provided by the authors.
Due to the design of these methods, the data preparation routines and the input signals are different for each dataset, and sometimes deviate from the guidelines provided in the papers.

\mypara{PointNet.} For indoor datasets, we used the data sampling strategy suggested in the original paper with global $xyz$, locally normalized $xyz$, and RGB as inputs.
For Semantic3D, we observed global $xyz$ to be harmful, thus we only use local $xyz$ and color.
Training data is generated by randomly sampling $(3\text{m})^3$ cubes from the training scenes.
Evaluation is performed by applying a sliding window over the entire scan.

\mypara{ScanNet.} The original network used 2 input channels: occupancy and a visibility mask computed using known camera trajectories. Since scenes in general are not accompanied by known camera trajectories, we only use occupancy in the input signal. Following the original setup, we use $1.5 \timess 1.5 \timess 3$m volumes voxelized into $31 \timess 31 \timess 62$ grids and augmented by 8 rotations.
Each such cube yields a prediction for one $1 \timess 1 \timess 62$ column. (I.e., the ScanNet network outputs a prediction for the central column only.) We use random sampling for training, and exhaustive sliding window for testing.

\mypara{OctNet.}
We use an architecture that operates on $256^3$ octrees.
Inputs to the network are color, occupancy, and a height-based feature that assigns each point to the top or bottom part of the scan.
Based on correspondence with the authors regarding the best way to set up OctNet on different datasets, we used $(45\text{m})^3$ volumes for Semantic3D and $(11\text{m})^3$ volumes for the indoor datasets.

\subsection{Setup of the presented approach}

The architecture described in Section~\ref{sec:architecture} is used in all experiments.
We evaluate four variants that use different input signals: distance from tangent plane (D), height above ground (H), normals (N), and color (RGB).
All input signals are normalized between 0 and 1.
The initial resolution $r$ of the tangent image is 5cm for the indoor datasets and 10cm for Semantic3D.
It is doubled after each pooling layer.
In addition to providing the distance from tangent plane as input to the first convolutional layer, we concatenate the local distance features from all scales of the point cloud to the feature maps of the corresponding resolution produced by pooling layers.

For ScanNet and S3DIS, we used whole rooms as individual training batches.
For Semantic3D, each batch was a random sphere with a radius of 6m.
For indoor scans, we augment each scan by 8 rotations around the vertical axis.
To correct for imbalance between different classes, we weigh the loss with the negative log of the training data histogram.

\subsection{Signal interpolation}

We begin by comparing the effectiveness of two different signal interpolation schemes: nearest neighbor and Gaussian mixture.
Both networks were trained on S3DIS with D and H as the input signals.
The resulting segmentation scores are provided in the supplement.
The two networks produce similar results.
We conclude that the nearest neighbor signal estimation scheme is sufficient, and use it in all other experiments.

\subsection{Main results}

\begin{figure*}
\centering
    
\definecolor{stanford_1}{rgb}{0.50196078,  0.50196078,  0.50196078}
\definecolor{stanford_2}{rgb}{0.48627451,  0.59607843,  0.        }
\definecolor{stanford_3}{rgb}{0.76470588,  0.64705882,  0.09803922}
\definecolor{stanford_4}{rgb}{0.        ,  0.50980392,  0.78431373}
\definecolor{stanford_5}{rgb}{0.2745098 ,  0.94117647,  0.94117647}
\definecolor{stanford_6}{rgb}{0.56862745,  0.11764706,  0.70588235}
\definecolor{stanford_7}{rgb}{0.        ,  0.50980392,  0.78431373}
\definecolor{stanford_8}{rgb}{0.        ,  0.        ,  0.50196078}
\definecolor{stanford_9}{rgb}{0.50196078,  0.        ,  0.        }
\definecolor{stanford_10}{rgb}{0.56862745,  0.11764706,  0.70588235}
\definecolor{stanford_11}{rgb}{0.66666667,  0.43137255,  0.15686275}
\definecolor{stanford_12}{rgb}{0.        ,  0.        ,  0.        }
\definecolor{stanford_13}{rgb}{0.90196078,  0.74509804,  1.        }
\definecolor{stanford_14}{rgb}{0.50196078,  0.50196078,  0.50196078}

\newcommand{\wind}{63}
\newcommand{\hind}{30.0}

\resizebox*{0.92\textwidth}{!}{
\begin{tabular}{@{}c@{\hspace{1mm}}c@{\hspace{1mm}}c@{\hspace{1mm}}}

	\includegraphics[width={\wind}mm, totalheight={\hind}mm, trim={0 6mm 0 44mm}, clip]{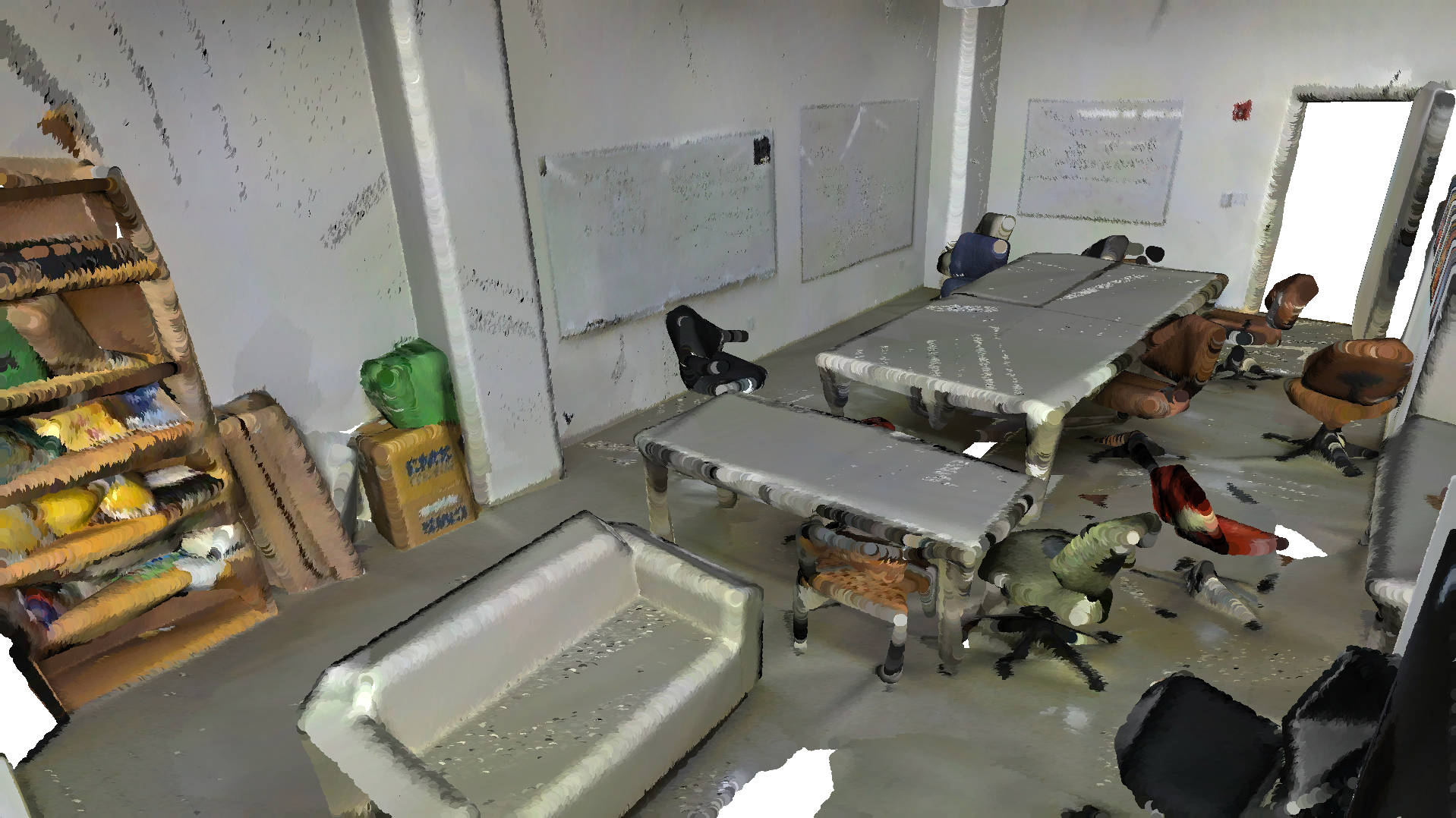}
    & \includegraphics[width={\wind}mm, totalheight={\hind}mm, trim={0 6mm 0 44mm}, clip]{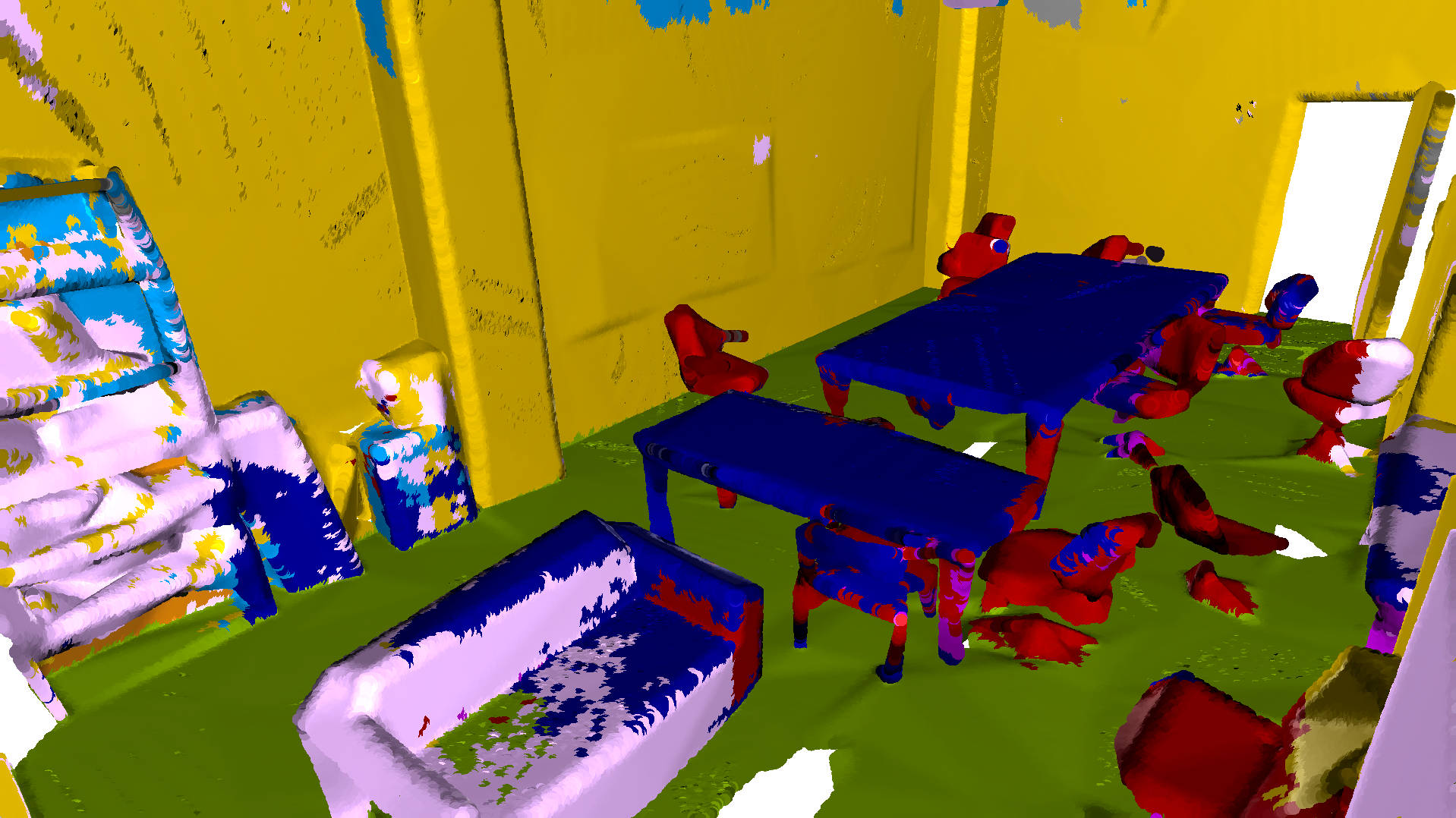}
    & \includegraphics[width={\wind}mm, totalheight={\hind}mm, trim={0 6mm 0 44mm}, clip]{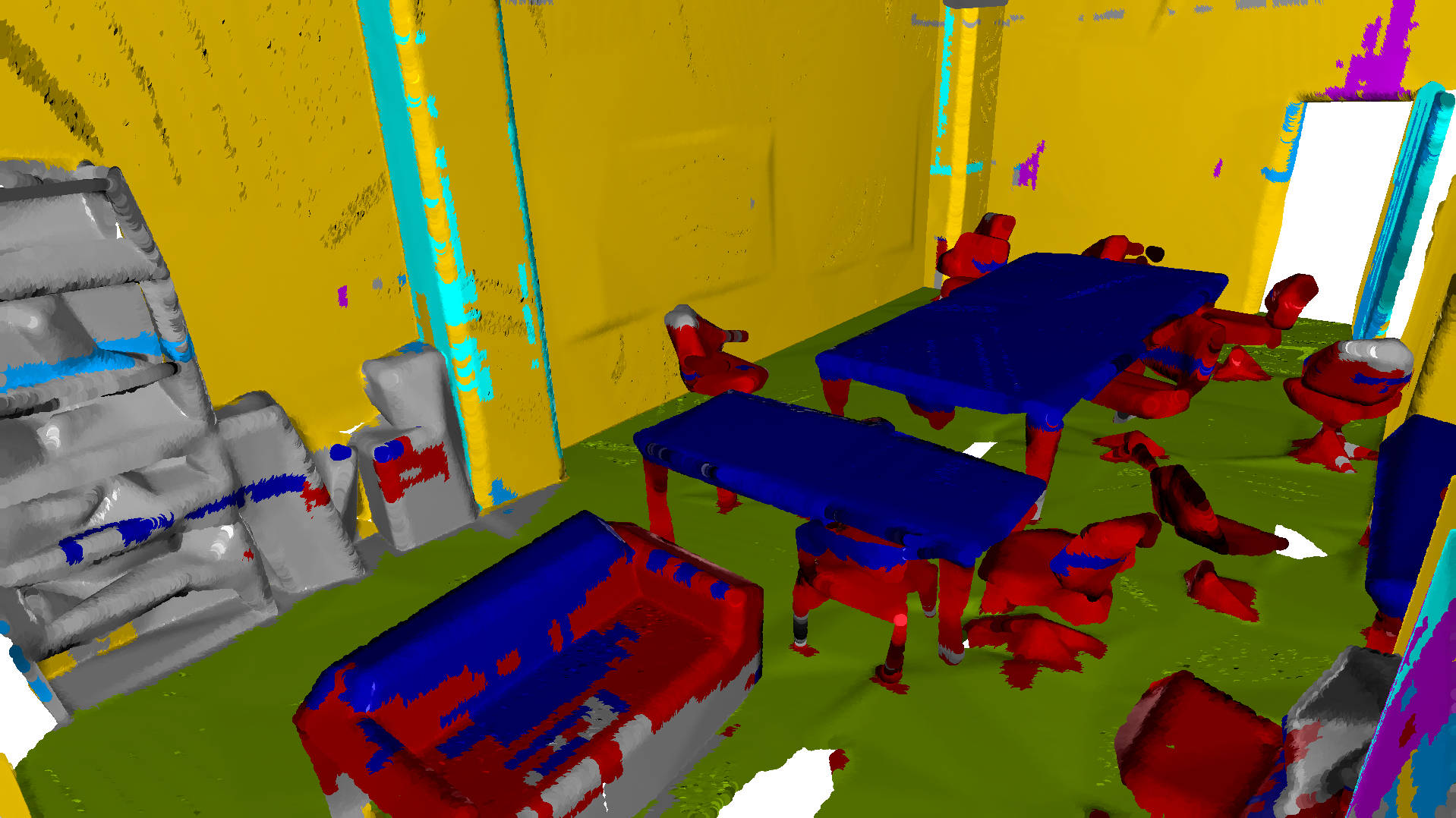} \\
	Color & PointNet~\cite{qi17} & ScanNet~\cite{dai17}\\
	\includegraphics[width={\wind}mm, totalheight={\hind}mm, trim={0 6mm 0 44mm}, clip]{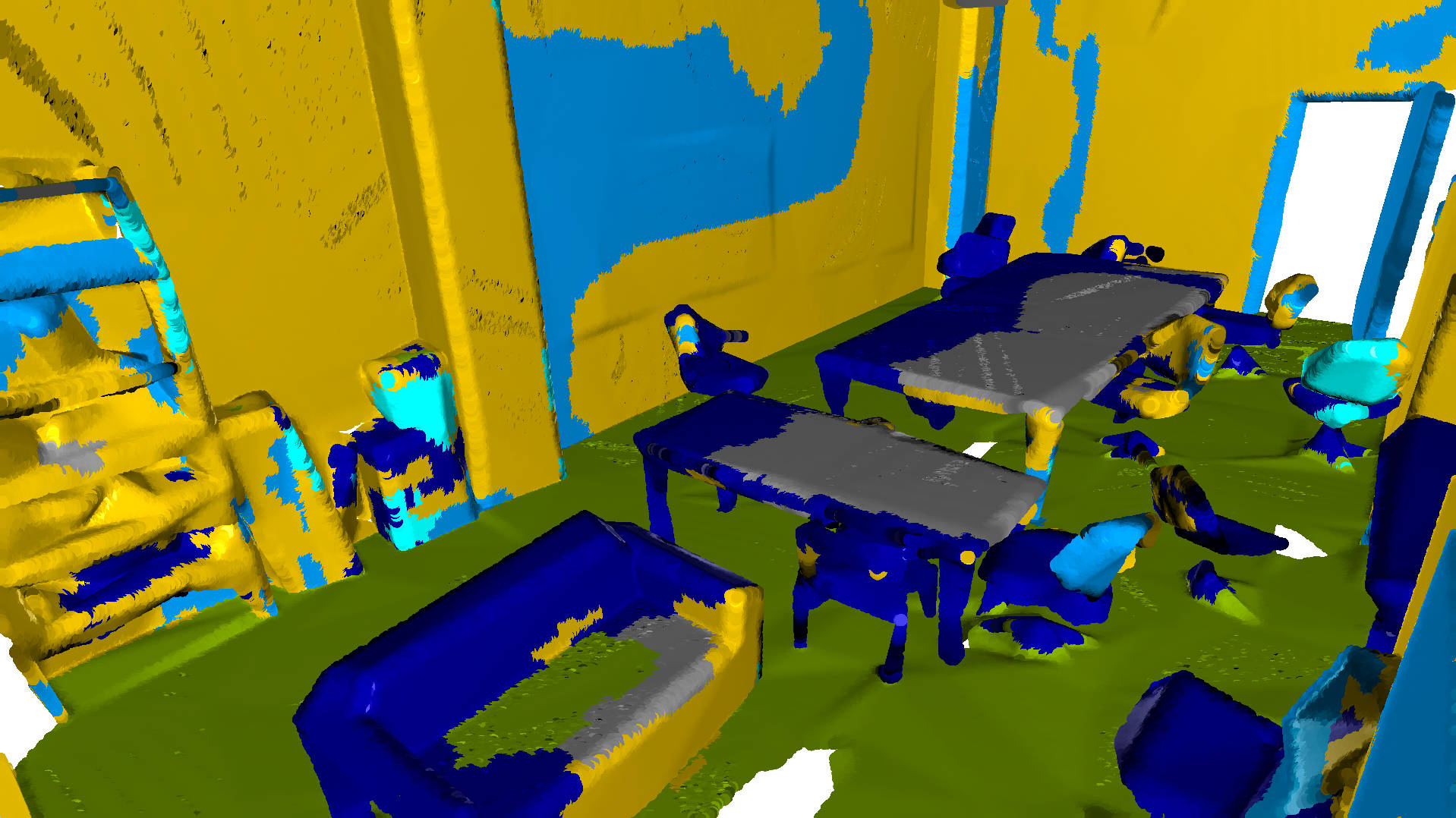}
    & \includegraphics[width={\wind}mm, totalheight={\hind}mm, trim={0 6mm 0 44mm}, clip]{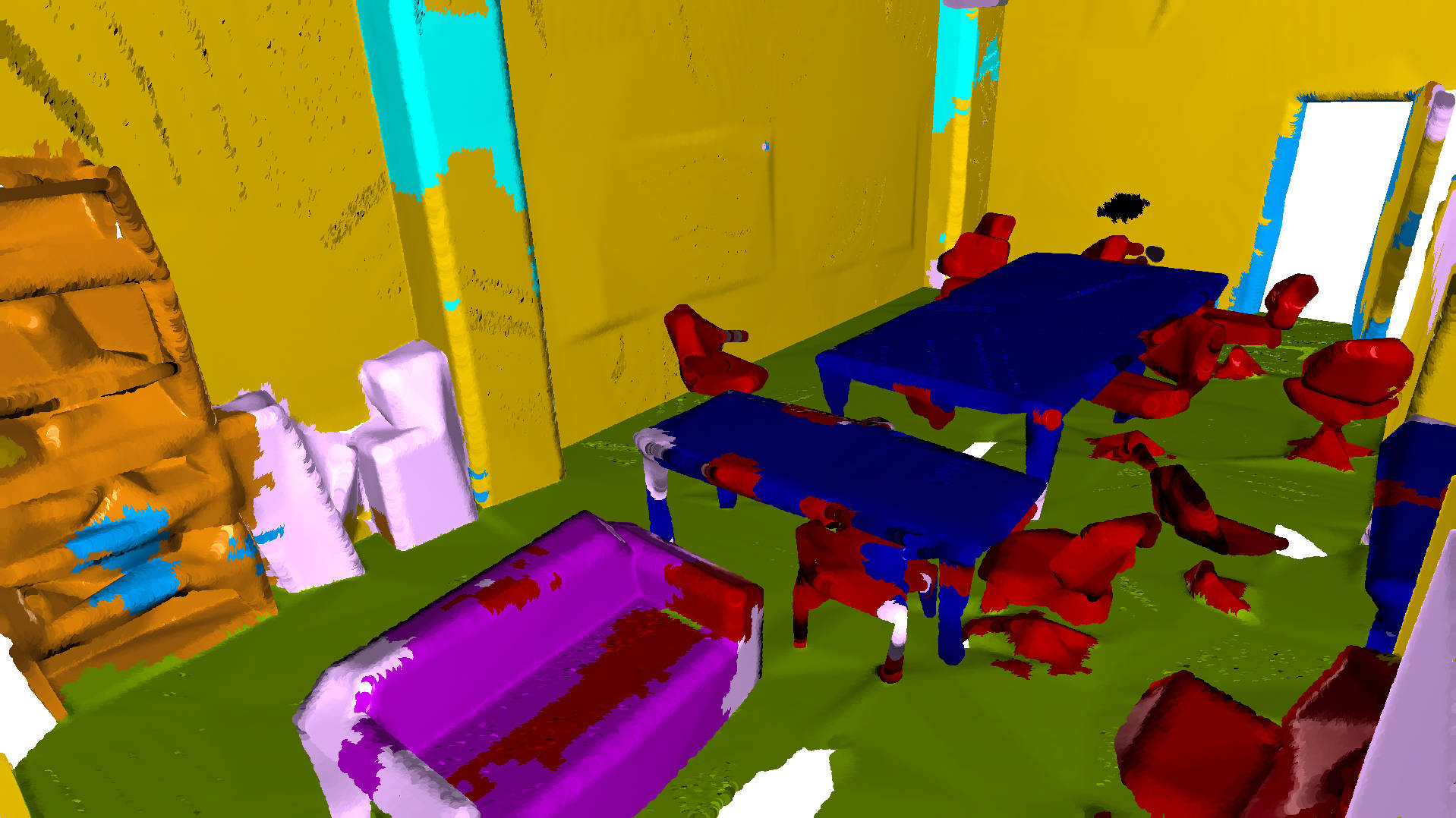}
    & \includegraphics[width={\wind}mm, totalheight={\hind}mm, trim={0 6mm 0 44mm}, clip]{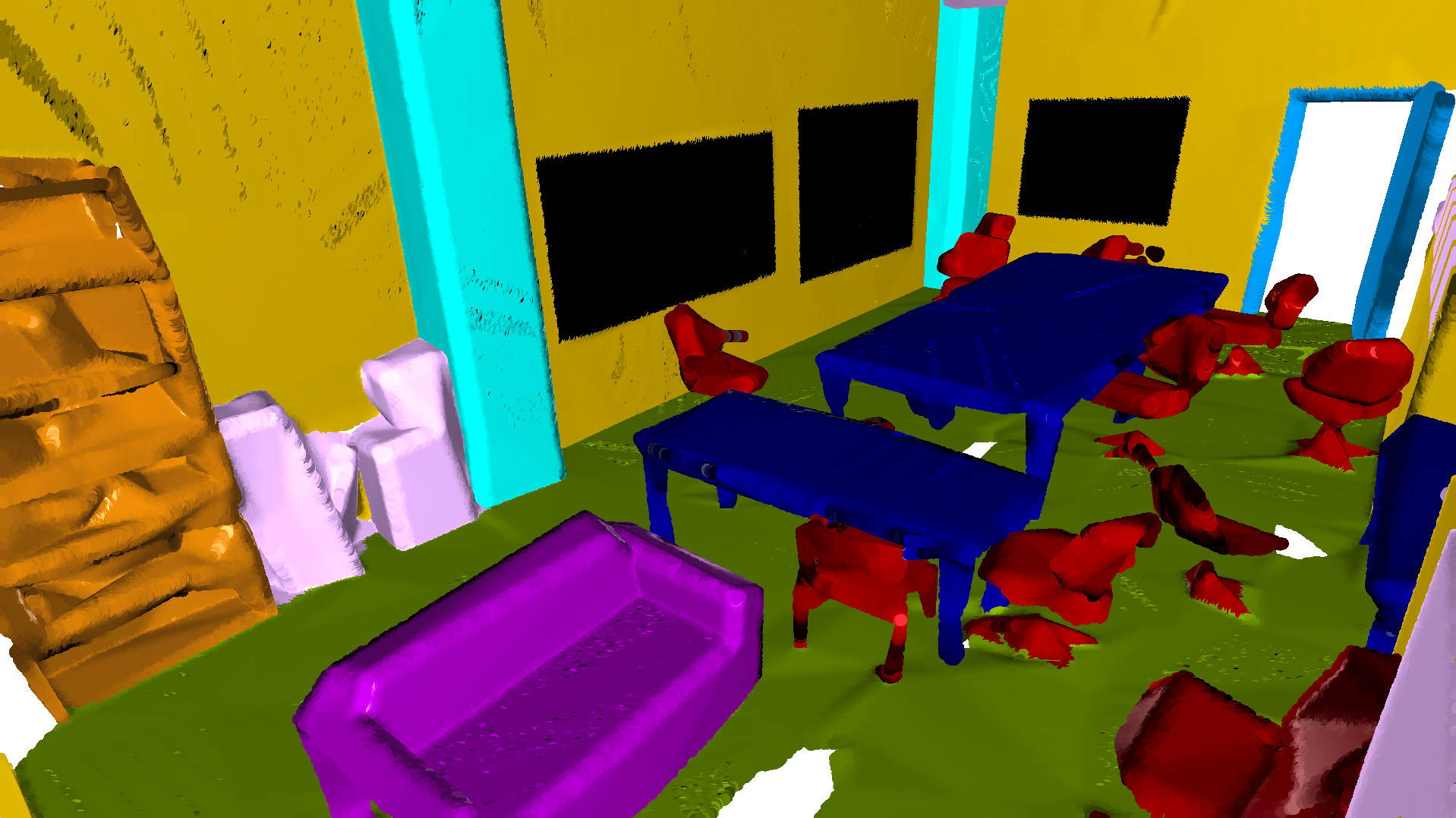} \\
	OctNet~\cite{riegler17} & Ours (DHNRGB) & Ground truth\\
	\\
	\multicolumn{3}{c}{
		\textcolor{stanford_1}{\ColorMapCircle} Ceiling
		\textcolor{stanford_2}{\ColorMapCircle} Floor
		\textcolor{stanford_3}{\ColorMapCircle} Walls
		\textcolor{stanford_5}{\ColorMapCircle} Column
		\textcolor{stanford_7}{\ColorMapCircle} Door
		\textcolor{stanford_8}{\ColorMapCircle} Table
		\textcolor{stanford_9}{\ColorMapCircle} Chair
		\textcolor{stanford_10}{\ColorMapCircle} Sofa
		\textcolor{stanford_11}{\ColorMapCircle} Bookcase
		\textcolor{stanford_12}{\ColorMapCircle} Board
		\textcolor{stanford_13}{\ColorMapCircle} Clutter
	}\\\\
\end{tabular}
}

\definecolor{sem8_1}{rgb}{0.50196078,  0.50196078,  0.50196078}
\definecolor{sem8_2}{rgb}{1.        ,  0.88235294,  0.09803922}
\definecolor{sem8_3}{rgb}{0.48627451,  0.59607843,  0.        }
\definecolor{sem8_4}{rgb}{0.2745098 ,  0.94117647,  0.94117647}
\definecolor{sem8_5}{rgb}{0.50196078,  0.        ,  0.        }
\definecolor{sem8_6}{rgb}{0.96078431,  0.50980392,  0.18823529}
\definecolor{sem8_7}{rgb}{0.98039216,  0.74509804,  0.74509804}
\definecolor{sem8_8}{rgb}{0.        ,  0.50980392,  0.78431373}

\newcommand{\wout}{95}
\newcommand{\hout}{39.0}

\resizebox*{0.92\textwidth}{!}{
\begin{tabular}{@{}c@{\hspace{1mm}}c@{\hspace{1mm}}c}

	\includegraphics[width={\wout}mm, totalheight={\hout}mm, trim={0 7mm 0 7mm}, clip]{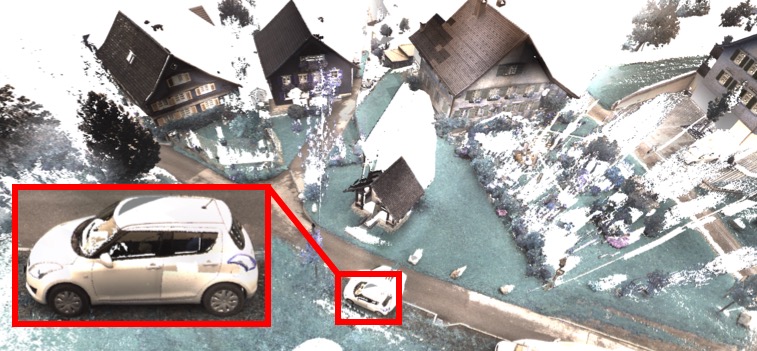}
    & \includegraphics[width={\wout}mm, totalheight={\hout}mm, trim={0 7mm 0 7mm}, clip]{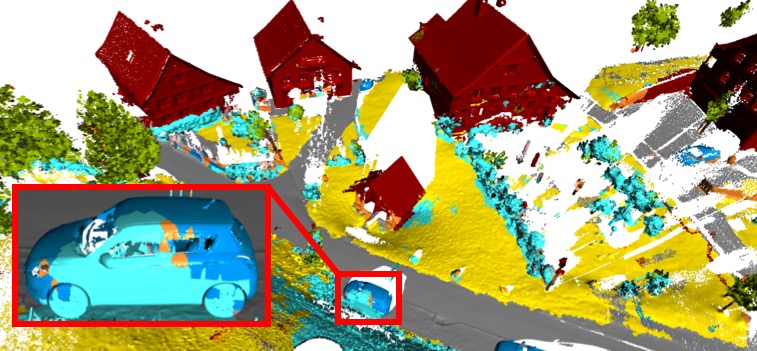} \\
	Color & OctNet~\cite{riegler17}\\
    \includegraphics[width={\wout}mm, totalheight={\hout}mm, trim={0 7mm 0 7mm}, clip]{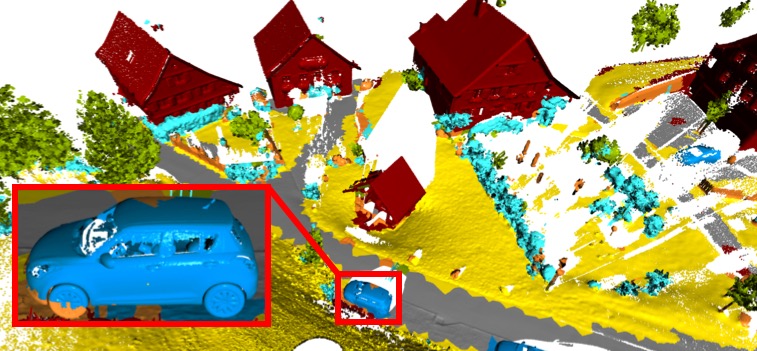}
    & \includegraphics[width={\wout}mm, totalheight={\hout}mm, trim={0 7mm 0 7mm}, clip]{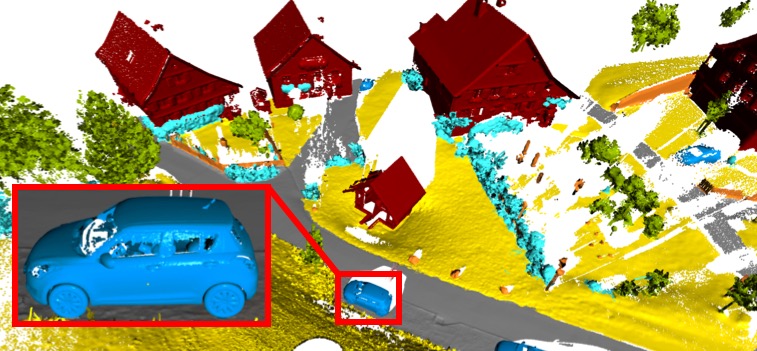} \\
	Ours (DHNRGB) & Ground truth\\
	\\
	\multicolumn{3}{c}{
		\textcolor{sem8_1}{\ColorMapCircle} Man made terrain
		\textcolor{sem8_2}{\ColorMapCircle} Natural terrain
		\textcolor{sem8_3}{\ColorMapCircle} High vegetation
		\textcolor{sem8_4}{\ColorMapCircle} Low vegetation
		\textcolor{sem8_5}{\ColorMapCircle} Building
		\textcolor{sem8_6}{\ColorMapCircle} Hardscape
		\textcolor{sem8_7}{\ColorMapCircle} Scanning artifacts
		\textcolor{sem8_8}{\ColorMapCircle} Cars
	}\\\\
\end{tabular}
}

\vspace{-1.5mm}
\caption{Qualitative comparisons on S3DIS \cite{armeni16} (top) and Semantic3D~\cite{hackel17} (bottom). Labels are coded by color.}
\label{fig:results}
\end{figure*}

Quantitative results for all methods are summarized in Table~\ref{tbl:results}.
Overall, our method produces high scores on all datasets and consistently outperforms the baselines.
Qualitative comparisons are shown in Figure~\ref{fig:results}.

Comparing the configurations of our networks that use different input signals, we can see that geometry is much more important than color on the indoor datasets.
Adding RGB information only slightly improves the scores on S3DIS and is actually harmful for mean and overall accuracy on the ScanNet dataset.
The situation is different for the Semantic3D dataset: the network trained with color significantly outperforms all other configurations.
Due to the fact that H is normalized between 0 and 1 for every scan separately, this information turns out to be harmful when the global height of different scans is significantly different.
Therefore, the network trained only with the distance signal performs better than the other two geometric configurations.


In setting up and operating the baseline methods, we found that all of them are quite hard to apply across datasets: some non-trivial decisions had to be made for each new dataset during the data preparation stage.
None of the baselines showed consistent performance across the different types of scenes.

PointNet reaches high oA scores on both indoor datasets. However, the oA measure is strongly dominated by large classes such as walls, floor, and ceiling.
S3DIS has a fairly regular layout because of the global room alignment procedure, which is very beneficial for PointNet and allows it to reach reasonable mA and mIoU scores on this dataset.
However, PointNet performs poorly on the ScanNet dataset, which has more classes and noisy data.
All but the most prominent classes (i.e., walls and floor) are misclassified.
PointNet completely fails to produce meaningful predictions on the even more challenging Semantic3D dataset.

Our configuration of the ScanNet method produces reasonable oA scores on both indoor datasets, but does much worse in the other two measures.
For reference, on the ScanNet dataset we additionally report the number from the original paper where a binary visibility-from-camera mask was used as an additional input channel.
This number is much higher than our occupancy-only results, which do not assume a known camera trajectory. Due to the fact that the network only outputs predictions for the central column of the voxel grid, evaluation is extremely time-consuming for the large scenes in the Semantic3D dataset.
Because of this scalability issue, we did not succeed in evaluating ScanNet on this dataset.

OctNet reaches good performance on the Semantic3D dataset.
However, the same network configuration yields bad results when applied to the indoor datasets.
A possible explanation for this may be poor generalization due to overfitting to the structure of training octrees.

\subsection{Efficiency}

We compared the efficiency of different methods on a scan from S3DIS containing 125K points after grid hashing. The results are reported in Table~\ref{tbl:runtime}.
Since ScanNet and PointNet require multiple iterations for labeling a single scan, we report both the time of a single forward pass and the time for processing a full scan.
OctNet and our method process a full scan in one forward pass, which also explains their higher memory consumption compared to ScanNet and PointNet.
ScanNet does not provide code for data preprocessing, so we report the runtime of our Python implementation needed for generating 38K sliding windows during inference.
Our method exhibits the best runtime for both precomputation and inference.

\begin{table}[h]
\small
\centering
	\setlength{\tabcolsep}{2mm}
	\ra{0.9}
	\begin{tabular}{@{}l c c c c}
	\toprule
	& Prep (s) & FP (s) & Full (s) & Mem (GB)\\
	\midrule
	PointNet & 16.5 & 0.01 & 0.65 & 0.39\\
	OctNet & 15.5 & 0.61 & 0.61 & 3.33\\
	ScanNet & 867.8 & 0.002 & 6.34 & 0.97\\
	\midrule
	Ours & 1.59 & 0.52 & 0.52 & 2.35\\
	\bottomrule
	\end{tabular}
\vspace{1mm}
\caption{Efficiency of different methods. We report preprocessing time (Prep), time for a single forward pass (FP), time for processing a full scan (Full), and memory consumption (Mem).}
\label{tbl:runtime}
\vspace{-2mm}
\end{table}

\begin{table*}[ht!]
\centering
\resizebox*{0.95\textwidth}{!}{
	\setlength{\tabcolsep}{4mm}
	\ra{1}
	\begin{tabular}{@{}l c c c c c c c c c}
	\toprule
	 & \multicolumn{3}{c}{Semantic3D~\cite{hackel17}} & \multicolumn{3}{c}{ScanNet~\cite{dai17}} & \multicolumn{3}{c}{S3DIS~\cite{armeni16}}\\
	\cmidrule(l{3mm}r{3mm}){2-4} \cmidrule(l{3mm}r{3mm}){5-7} \cmidrule(l{3mm}r{3mm}){8-10}
	& mIoU & mA & oA & mIoU & mA & oA & mIoU & mA & oA\\
	\midrule
	PointNet~\cite{qi17} & 3.76 & 16.9 & 16.3 & 12.2 & 17.9 & 68.1 & 41.3 & 49.5 & 78.8\\
	OctNet~\cite{riegler17} & 50.7 & 71.3 & 80.7 & 18.1 & 26.4 & 76.6 & 26.3 & 39.0 & 68.9\\
	ScanNet~\cite{dai17} & n/a & n/a & n/a & 13.5 & 19.2 (\textit{50.8}) & 69.4 (\textit{73.0}) & 24.6 & 35.0 & 64.2\\
	\midrule
	Ours (D) & 58.1 & 78.9 & 84.8 & 40.9 & 52.5 & \textbf{80.9} & 49.8 & 60.3 & 80.2\\
	Ours (DH) & 58.0 & 75.8 & 83.3 & 40.3 & 52.2 & 80.6 & 50.0 & 60.0 & 81.2\\
	Ours (DHN) & 52.5 & 79.3 & 79.5 & 40.7 & \textbf{55.3} & 80.3 & 51.7 & 61.0 & 82.2\\
	Ours (DHNRGB) & \textbf{66.4} & \textbf{80.7} & \textbf{89.3} & \textbf{40.9} & 55.1 & 80.1 & \textbf{52.8} & \textbf{62.2} & \textbf{82.5}\\
	\bottomrule
	\end{tabular}
}
\vspace{1mm}
\caption{Semantic segmentation accuracy for all methods across the three datasets. We report mean intersection over union (mIoU), mean class accuracy (mA), and overall accuracy (oA). Note that oA is a bad measure and we recommend against using it in the future. We tested different configurations of our method by combining four types of input signals: depth (D), height (H), normals (N), and color (RGB).}
\label{tbl:results}
\vspace{3mm}
\end{table*}

\section{Conclusion}
\label{sec:discussion}
We have presented  tangent convolutions -- a new construction for convolutional networks on 3D data. The key idea is to evaluate convolutions on virtual tangent planes at every point. Crucially, tangent planes can be precomputed and deep convolutional networks based on tangent convolutions can be evaluated efficiently on large point clouds. We have applied tangent convolutions to semantic segmentation of large indoor and outdoor scenes. The presented ideas may also be applicable to other problems in analysis, processing, and synthesis of 3D data.




\pagebreak

\balance

{\small
\bibliographystyle{ieee}
\bibliography{paper}
}

\pagebreak

\section*{Appendix}

\appendix

\nobalance

\section{Robustness to noise}
\label{sec:noise}
We evaluated the robustness of our approach to noise. Several instances of our network were trained on the S3DIS dataset perturbed with different amounts of additive Gaussian noise with standard deviation $\sigma$. The results are reported in Table~\ref{tbl:noise}. We selected small subsets of the data for training and testing (Area 1 for training and Area 5 for testing), which is why the final performance numbers are not compatible with those reported in the main paper.

\begin{table}[h]
\small
\centering
	\setlength{\tabcolsep}{2mm}
	\ra{0.9}
	\begin{tabular}{@{\hspace{1mm}}c|@{\hspace{1mm}}c@{\hspace{1mm}}c@{\hspace{1mm}}c@{\hspace{1mm}}c@{\hspace{1mm}}c@{\hspace{1mm}}}
	\toprule
	\rotatebox{90}{\parbox[t]{16mm}{\hspace*{\fill}\footnotesize Training data\hspace*{\fill}}}
	& \includegraphics[width=14mm,trim={8mm 0 8mm 0},clip]{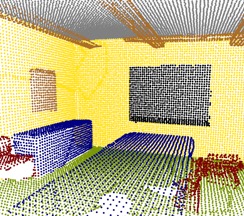}
	& \includegraphics[width=14mm,trim={8mm 0 8mm 0},clip]{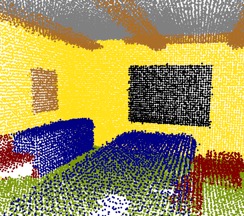}
	& \includegraphics[width=14mm,trim={8mm 0 8mm 0},clip]{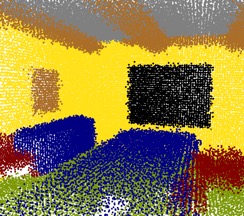}
	& \includegraphics[width=14mm,trim={8mm 0 8mm 0},clip]{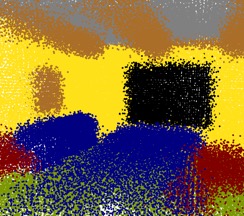}
	& \includegraphics[width=14mm,trim={8mm 0 8mm 0},clip]{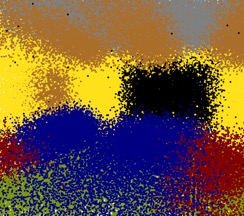}\\
	\midrule
	$\sigma$, m & 0.00 & 0.02 & 0.04 & 0.08 & 0.16\\
	\midrule
	OA & 0.59 & 0.63 & 0.63 & 0.68 & 0.17 \\
	\bottomrule
	\end{tabular}
\vspace{1mm}
\caption{Performance evaluation with different levels of noise.}
\label{tbl:noise}
\vspace{-2mm}
\end{table}

Surprisingly, reasonable amounts of noise improve overall accuracy. The method only suffers if the noise severely damages semantic structure in the point cloud.
We did not tune any parameters in the pipeline for these experiments.

\section{Signal interpolation}
\label{sec:interpolation}
In this experiment we compare the effectiveness of two signal interpolation schemes: nearest neighbor and Gaussian mixture.
Quantitative results on S3DIS using D and H as input signals are presented in Table \ref{tbl:sig_int}.
Both methods produce very similar results which is why the simpler nearest neighbor interpolation is used throughout the paper.

\begin{table}[H]
\small
\centering
	\setlength{\tabcolsep}{5mm}
	\ra{0.9}
	\begin{tabular}{@{}l c c c@{}}
	\toprule
	 Signal & mIoU & mA & oA\\
	\midrule
	 NN & 50.0 & 60.0 & 81.2\\
	 Gaussian & 50.7 & 59.6 & 81.3\\
	\bottomrule
	\end{tabular}
\vspace{1mm}
\caption{Signal interpolation using the nearest neighbor scheme and the Gaussian mixture scheme produce similar results.}
\label{tbl:sig_int}
\vspace{-2mm}
\end{table}

\section{Comparison with SnapNet}
\label{sec:snapnet}
We also compared our approach with the SnapNet by Boulch et al. \cite{boulch17}.
They project a 3D scene onto a set of 2D images.
Those images are then segmented with a regular 2D ConvNet.
The main strength of this approach is the possibility to combine it with transfer learning and use the weights of a network pre-trained on ImageNet for initialization.
Applying this strategy yields the mIoU score of 67.7 on the Semantic3D datset, compared to 66.4 produced by our approach.
However, the non-trivial camera pose sampling procedure required by SnapNet did not allow us to apply it to indoor datasets.

\section{Qualitative results}
\label{sec:qualitative}

\definecolor{stanford_1}{rgb}{0.50196, 0.50196, 0.50196}
\definecolor{stanford_2}{rgb}{0.48627, 0.59608, 0.0}
\definecolor{stanford_3}{rgb}{1.0, 0.88235, 0.09804}
\definecolor{stanford_4}{rgb}{0.0, 0.5098, 0.78431}
\definecolor{stanford_5}{rgb}{0.27059, 0.94118, 0.94118}
\definecolor{stanford_6}{rgb}{0.90196, 0.27059, 0.0}
\definecolor{stanford_7}{rgb}{0.0, 0.5098, 0.78431}
\definecolor{stanford_8}{rgb}{0.0, 0.0, 0.50196}
\definecolor{stanford_9}{rgb}{0.50196, 0.0, 0.0}
\definecolor{stanford_10}{rgb}{0.56078, 0.11765, 0.70588}
\definecolor{stanford_11}{rgb}{0.66667, 0.43137, 0.15686}
\definecolor{stanford_12}{rgb}{0.0, 0.0, 0.0}
\definecolor{stanford_13}{rgb}{0.90196, 0.7451, 1.0}
\definecolor{stanford_14}{rgb}{0.50196, 0.50196, 0.50196}

\definecolor{scannet_1}{rgb}{1.0, 0.88235, 0.09804}
\definecolor{scannet_2}{rgb}{0.48627, 0.59608, 0.0}
\definecolor{scannet_3}{rgb}{0.66667, 0.43137, 0.15686}
\definecolor{scannet_4}{rgb}{0.61961, 0.0, 0.55686}
\definecolor{scannet_5}{rgb}{0.50196, 0.0, 0.0}
\definecolor{scannet_6}{rgb}{0.56078, 0.11765, 0.70588}
\definecolor{scannet_7}{rgb}{0.0, 0.0, 0.50196}
\definecolor{scannet_8}{rgb}{0.0, 0.5098, 0.78431}
\definecolor{scannet_9}{rgb}{0.90196, 0.27059, 0.0}
\definecolor{scannet_10}{rgb}{0.66667, 0.43137, 0.15686}
\definecolor{scannet_11}{rgb}{0.0, 0.0, 0.0}
\definecolor{scannet_12}{rgb}{1.0, 0.89804, 0.00784}
\definecolor{scannet_13}{rgb}{0.50196, 0.50196, 0.50196}
\definecolor{scannet_14}{rgb}{1.0, 0.65098, 0.99608}
\definecolor{scannet_15}{rgb}{0.9098, 0.36863, 0.7451}
\definecolor{scannet_16}{rgb}{0.0, 0.39216, 0.00392}
\definecolor{scannet_17}{rgb}{0.52157, 0.66275, 0.0}
\definecolor{scannet_18}{rgb}{0.58431, 0.0, 0.22745}
\definecolor{scannet_19}{rgb}{0.0, 0.0, 1.0}
\definecolor{scannet_20}{rgb}{0.73333, 0.53333, 0.0}

\definecolor{sem8_1}{rgb}{0.50196078,  0.50196078,  0.50196078}
\definecolor{sem8_2}{rgb}{1.        ,  0.88235294,  0.09803922}
\definecolor{sem8_3}{rgb}{0.48627451,  0.59607843,  0.        }
\definecolor{sem8_4}{rgb}{0.4788 ,  0.43137,  0.15686}
\definecolor{sem8_5}{rgb}{0.50196078,  0.        ,  0.        }
\definecolor{sem8_6}{rgb}{0.96078431,  0.50980392,  0.18823529}
\definecolor{sem8_7}{rgb}{0.98039216,  0.74509804,  0.74509804}
\definecolor{sem8_8}{rgb}{0.        ,  0.50980392,  0.78431373}

\newcommand{\wind}{63}
\newcommand{\hind}{32.0}

\begin{figure*}
\centering
\resizebox*{1\textwidth}{!}{
\begin{tabular}{@{}c@{\hspace{1mm}}c@{\hspace{1mm}}c@{\hspace{1mm}}}
	
	Color & Prediction & Ground truth\\
	\includegraphics[width={\wind}mm, totalheight={\hind}mm, trim={0 20mm 0 20mm}, clip]{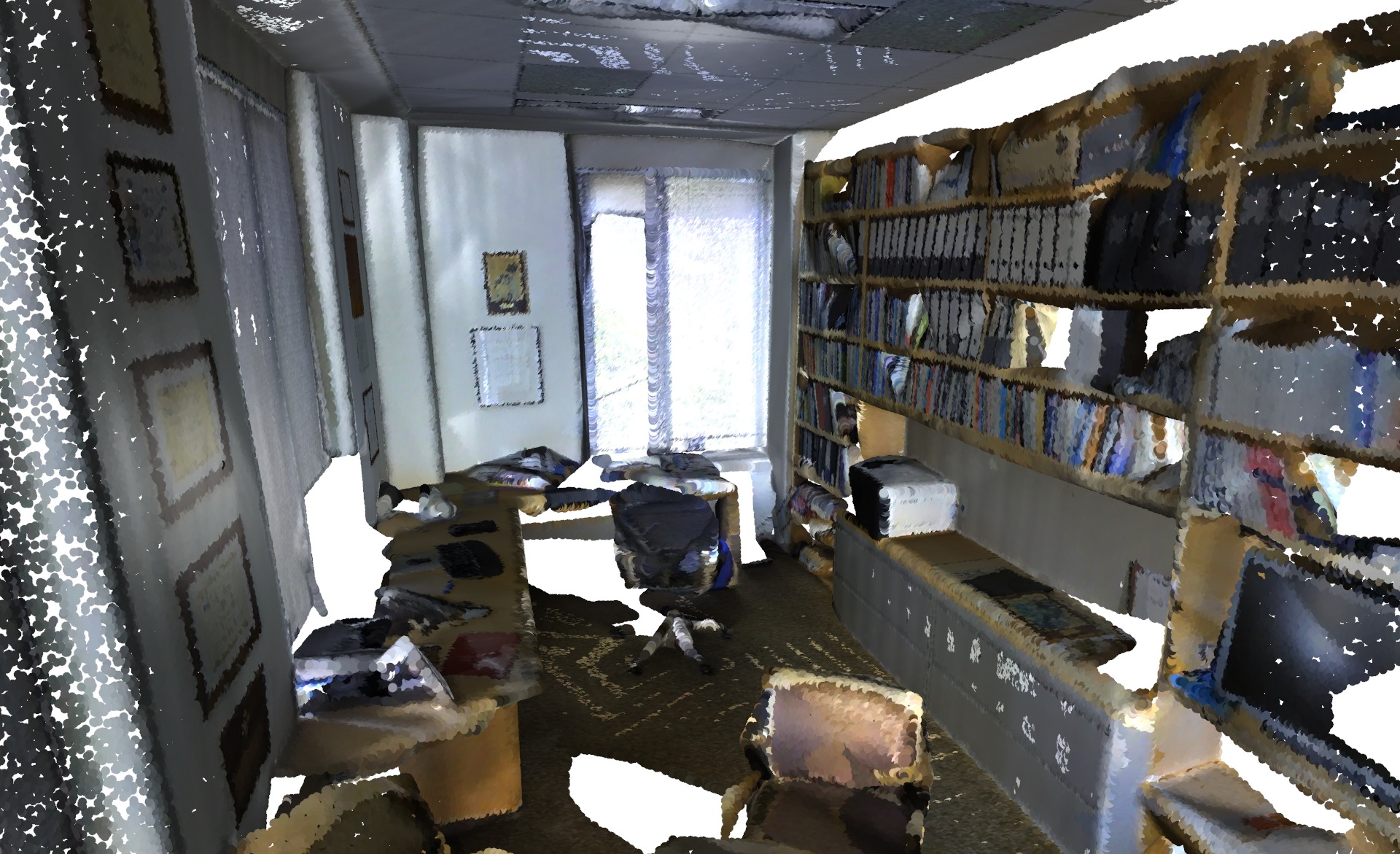}
    & \includegraphics[width={\wind}mm, totalheight={\hind}mm, trim={0 20mm 0 20mm}, clip]{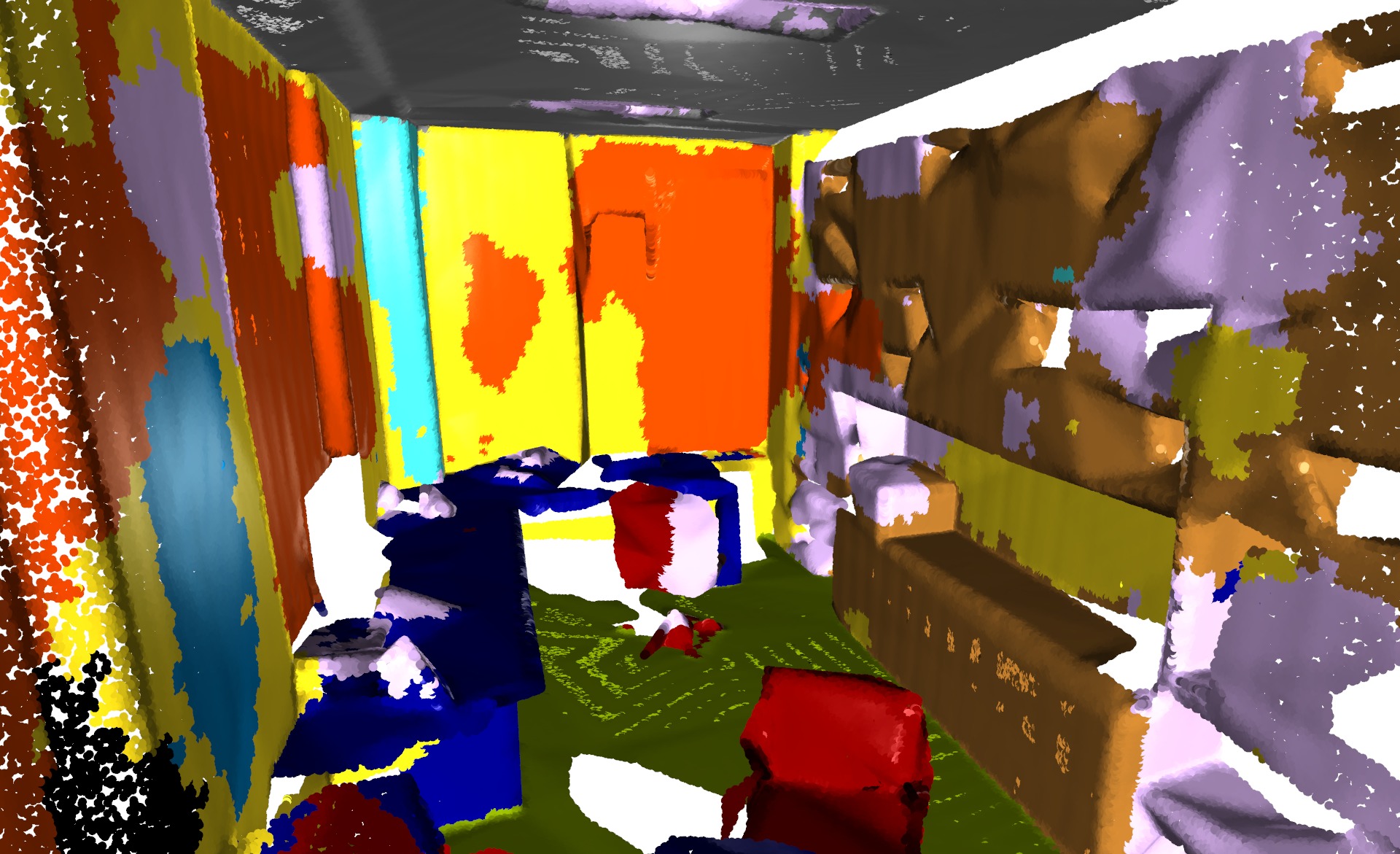}
    & \includegraphics[width={\wind}mm, totalheight={\hind}mm, trim={0 20mm 0 20mm}, clip]{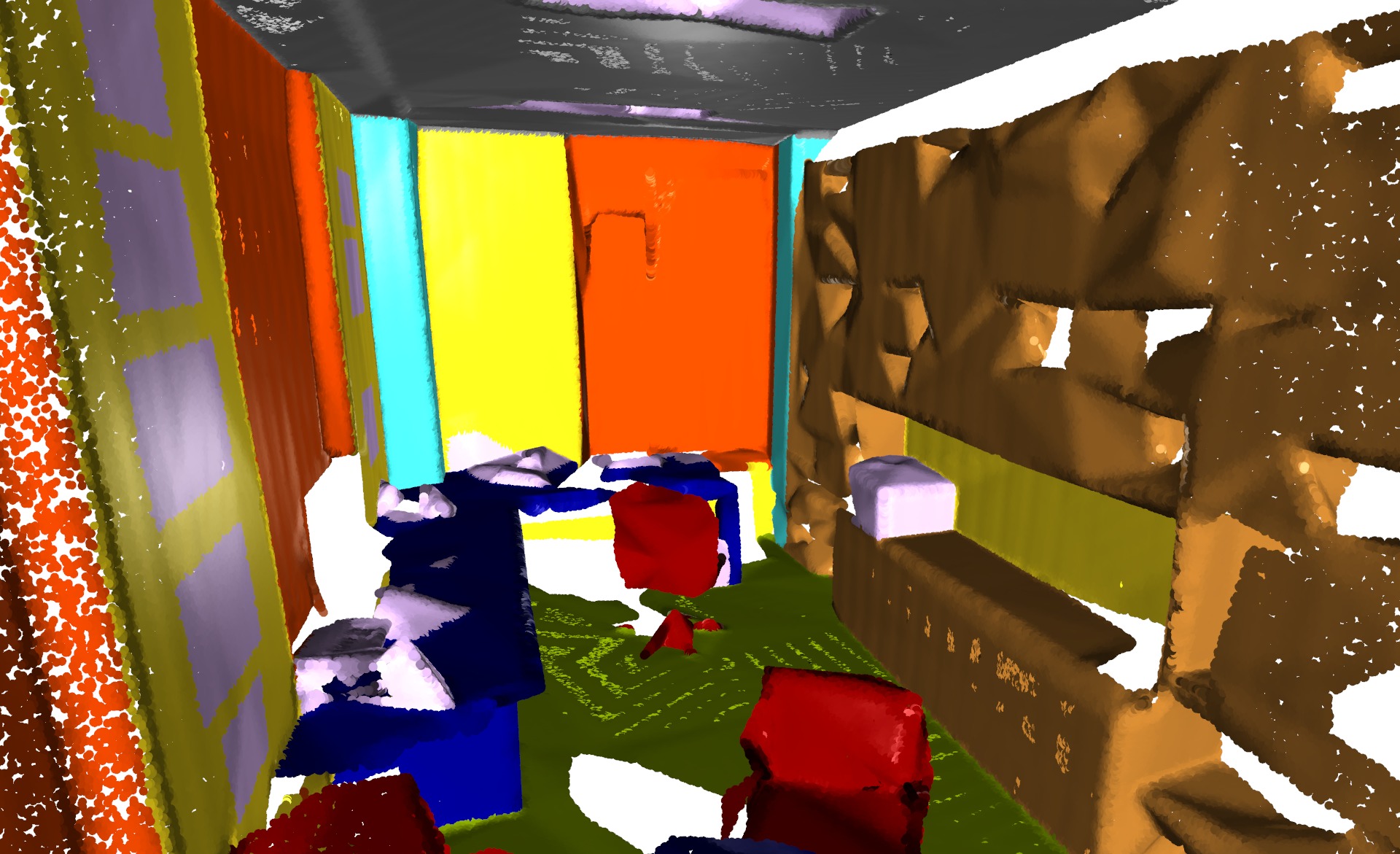} \\
    \includegraphics[width={\wind}mm, totalheight={\hind}mm, trim={0 20mm 0 20mm}, clip]{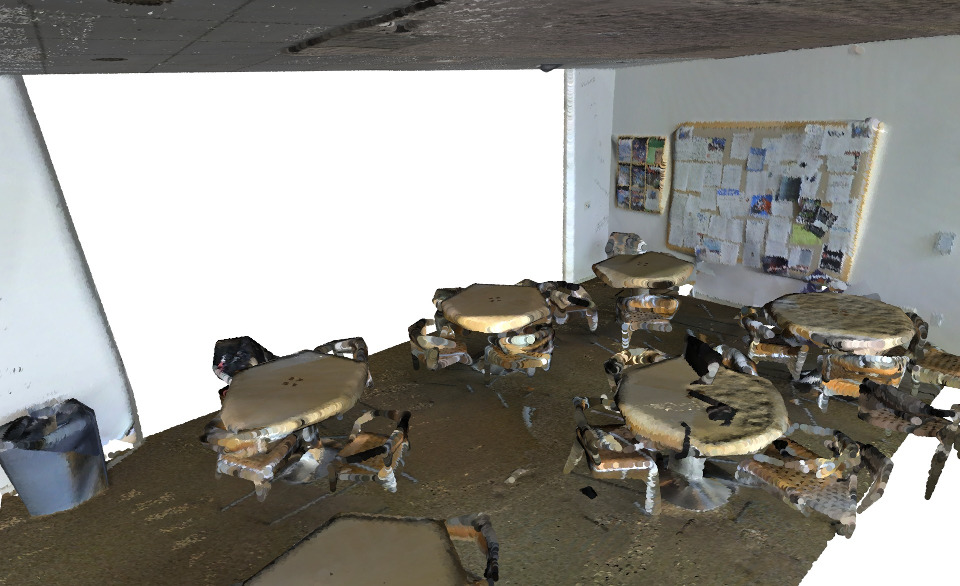}
    & \includegraphics[width={\wind}mm, totalheight={\hind}mm, trim={0 20mm 0 20mm}, clip]{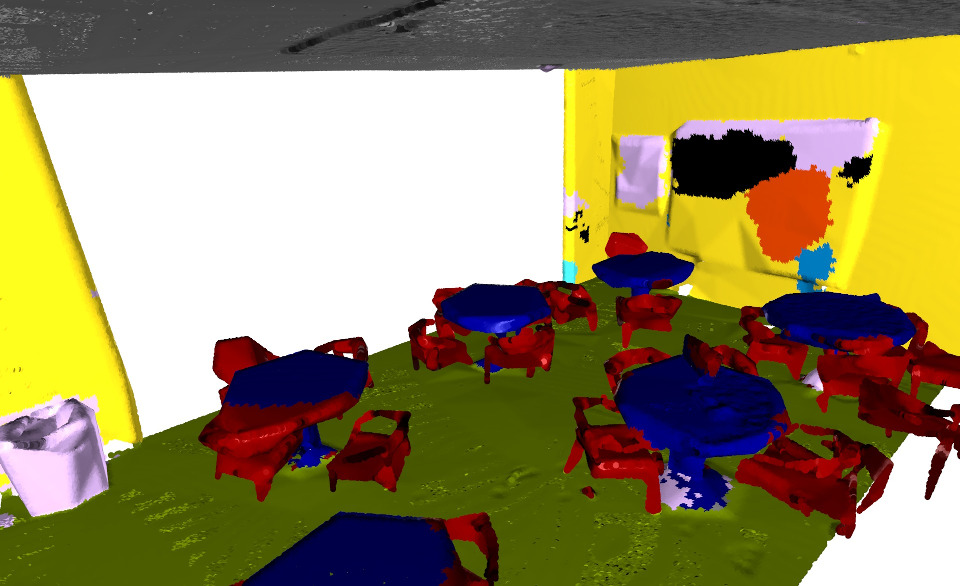}
    & \includegraphics[width={\wind}mm, totalheight={\hind}mm, trim={0 20mm 0 20mm}, clip]{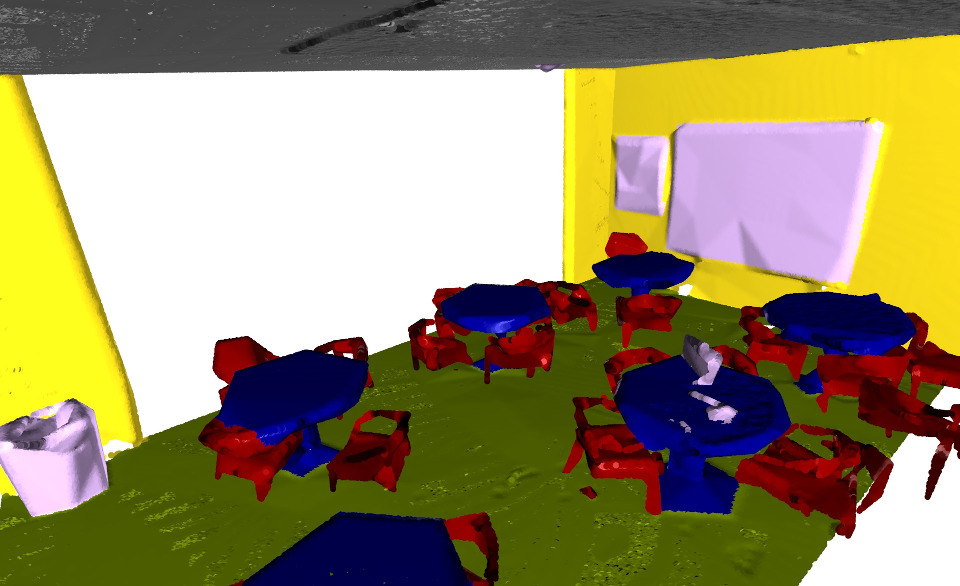} \\
    \includegraphics[width={\wind}mm, totalheight={\hind}mm, trim={0 20mm 0 20mm}, clip]{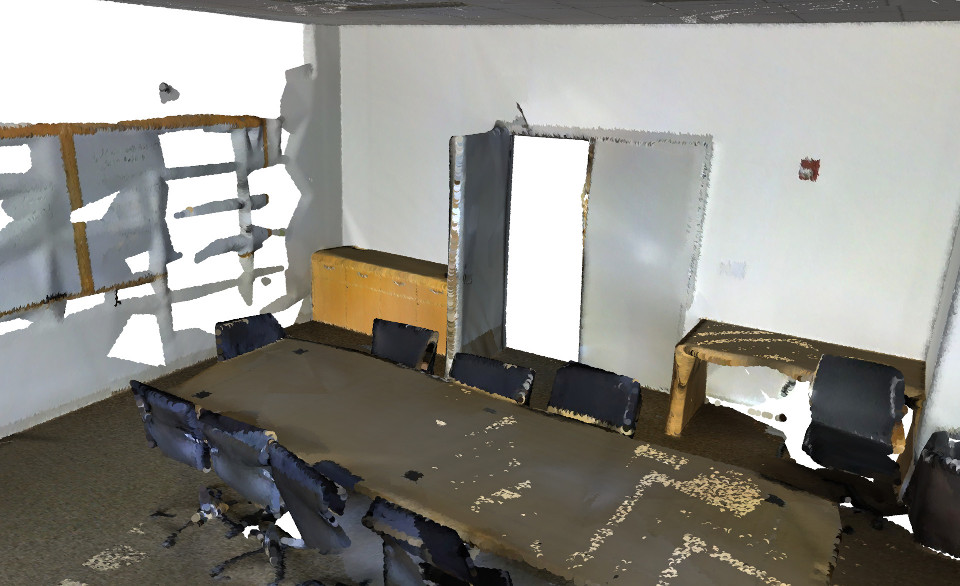}
    & \includegraphics[width={\wind}mm, totalheight={\hind}mm, trim={0 20mm 0 20mm}, clip]{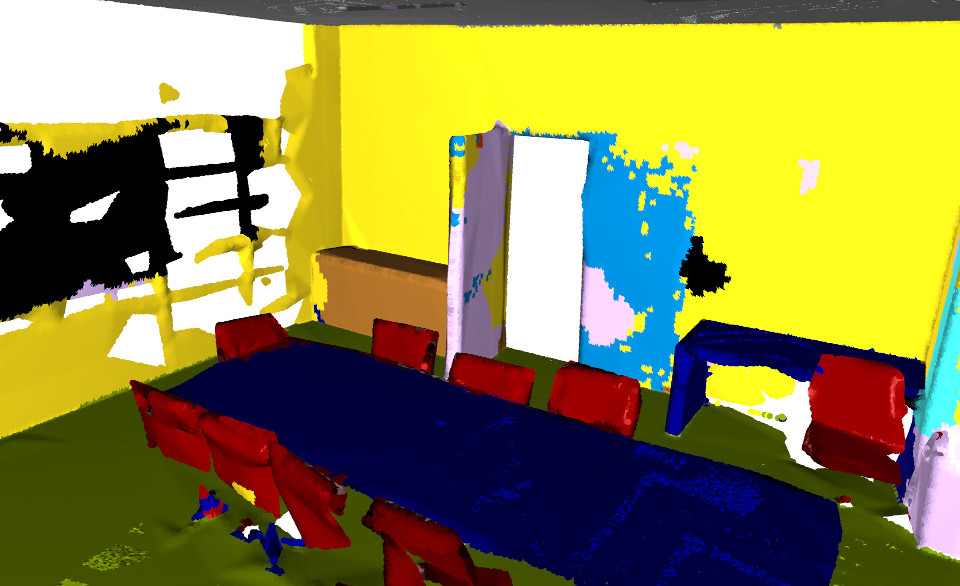}
    & \includegraphics[width={\wind}mm, totalheight={\hind}mm, trim={0 20mm 0 20mm}, clip]{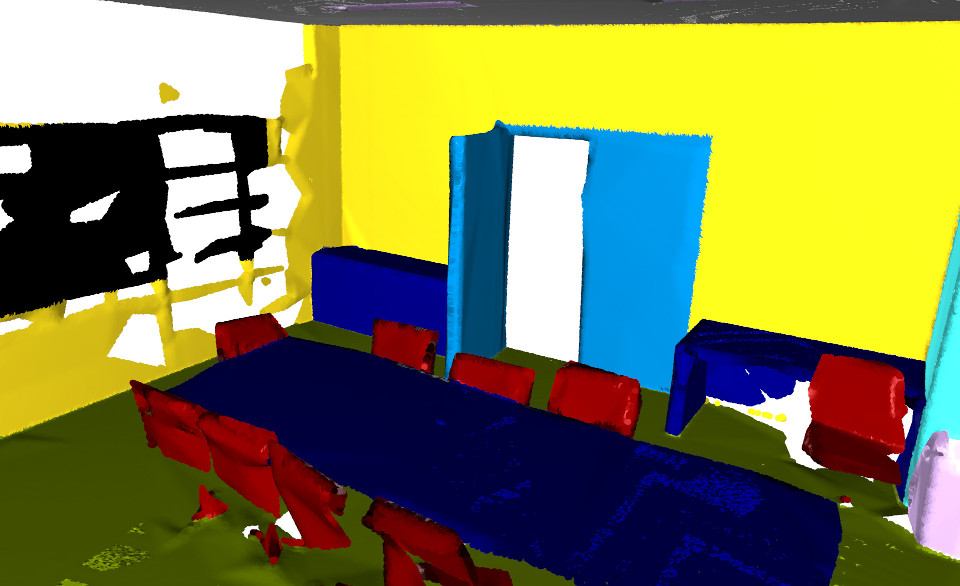} \\
	\multicolumn{3}{c}{
		\textcolor{stanford_1}{\ColorMapCircle} Ceiling
		\textcolor{stanford_2}{\ColorMapCircle} Floor
		\textcolor{stanford_3}{\ColorMapCircle} Walls
		\textcolor{stanford_5}{\ColorMapCircle} Column
		\textcolor{stanford_6}{\ColorMapCircle} Window
		\textcolor{stanford_7}{\ColorMapCircle} Door
		\textcolor{stanford_8}{\ColorMapCircle} Table
		\textcolor{stanford_9}{\ColorMapCircle} Chair
		\textcolor{stanford_10}{\ColorMapCircle} Sofa
		\textcolor{stanford_11}{\ColorMapCircle} Bookcase
		\textcolor{stanford_12}{\ColorMapCircle} Board
		\textcolor{stanford_13}{\ColorMapCircle} Clutter
	}\\\\
\end{tabular}
}
\caption{Qualitative results on S3DIS \cite{armeni16}.}
\label{fig:results_s3dis}
\end{figure*}

\begin{figure*}
\centering
\resizebox*{1\textwidth}{!}{
\begin{tabular}{@{}c@{\hspace{1mm}}c@{\hspace{1mm}}c@{\hspace{1mm}}}
	
	Color & Prediction & Ground truth\\
	\includegraphics[width={\wind}mm, totalheight={\hind}mm, trim={0 20mm 0 20mm}, clip]{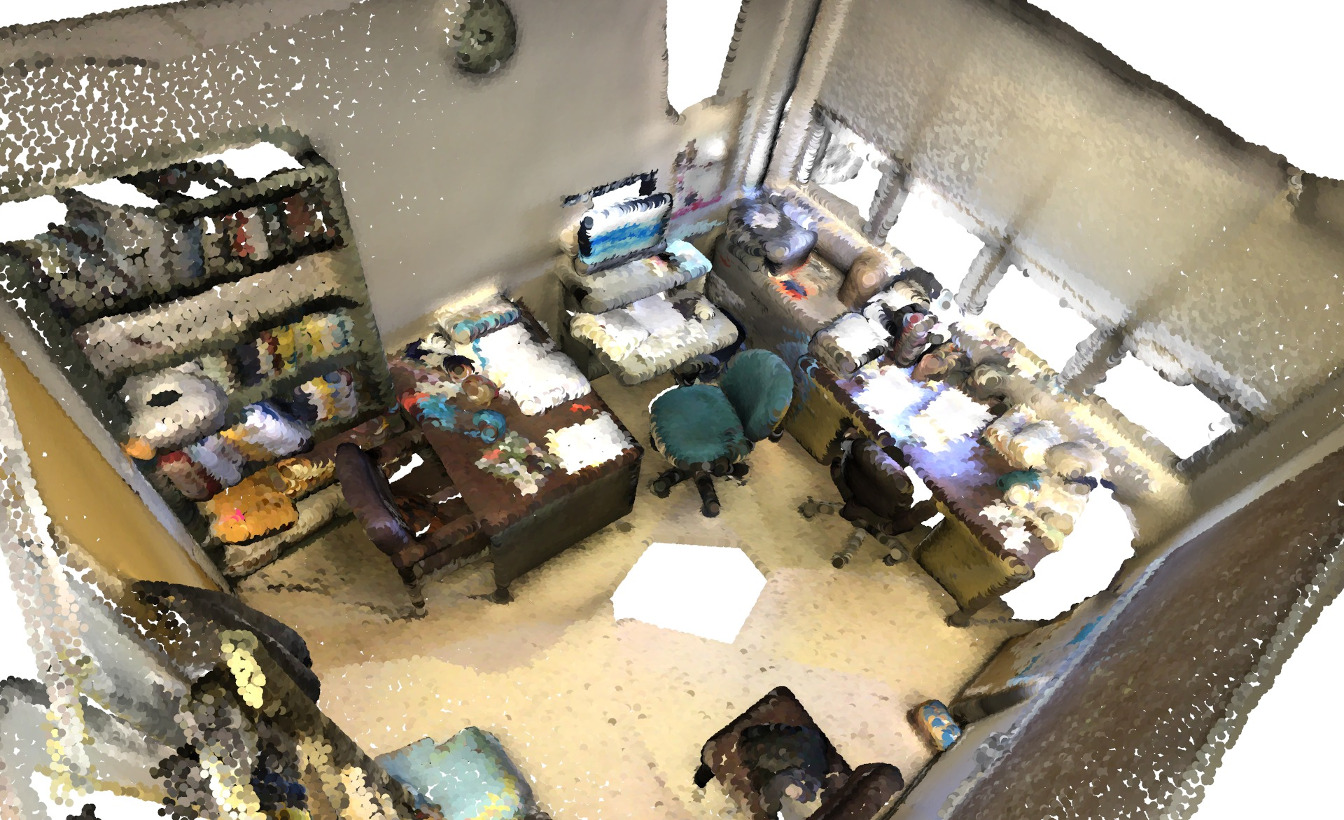}
    & \includegraphics[width={\wind}mm, totalheight={\hind}mm, trim={0 20mm 0 20mm}, clip]{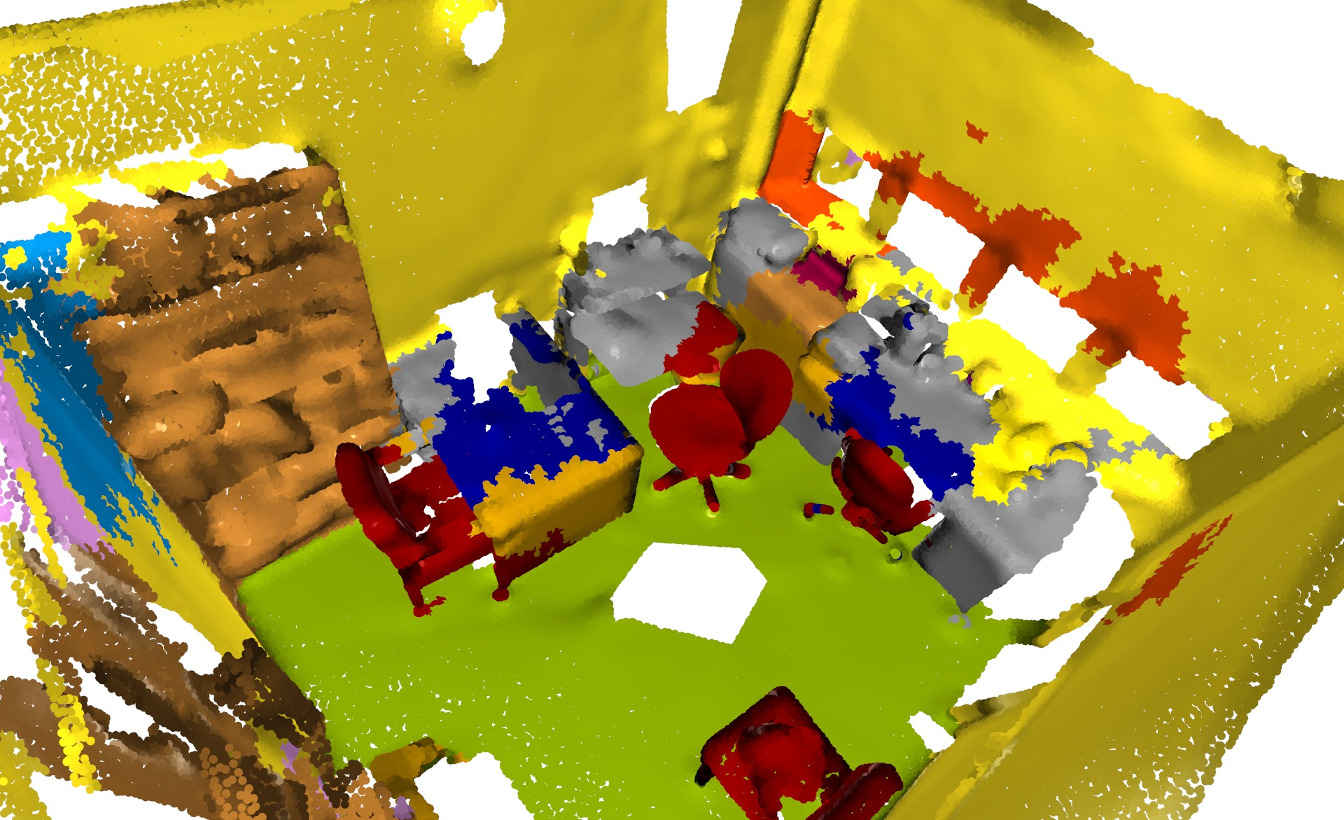}
    & \includegraphics[width={\wind}mm, totalheight={\hind}mm, trim={0 20mm 0 20mm}, clip]{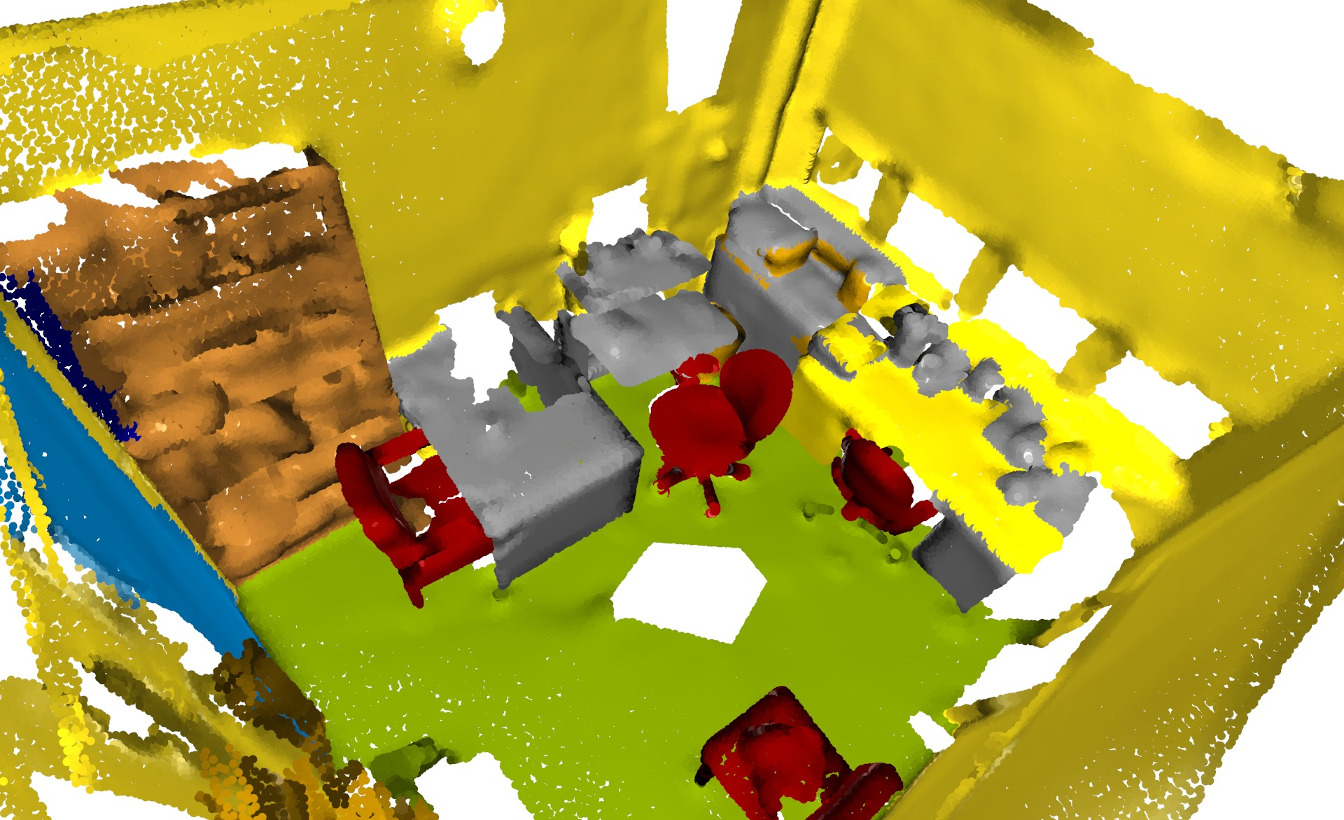} \\
    \includegraphics[width={\wind}mm, totalheight={\hind}mm, trim={0 20mm 0 20mm}, clip]{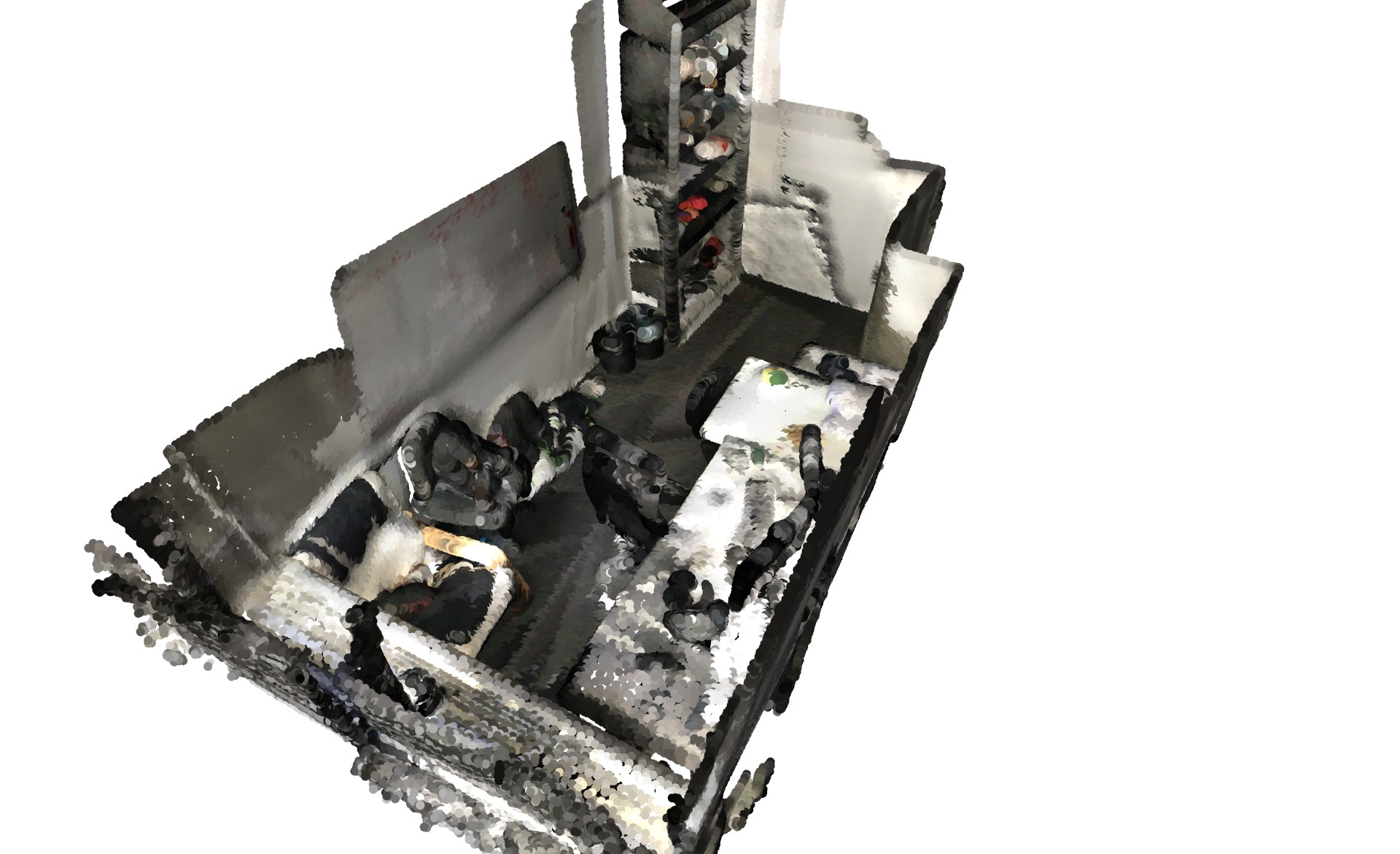}
    & \includegraphics[width={\wind}mm, totalheight={\hind}mm, trim={0 20mm 0 20mm}, clip]{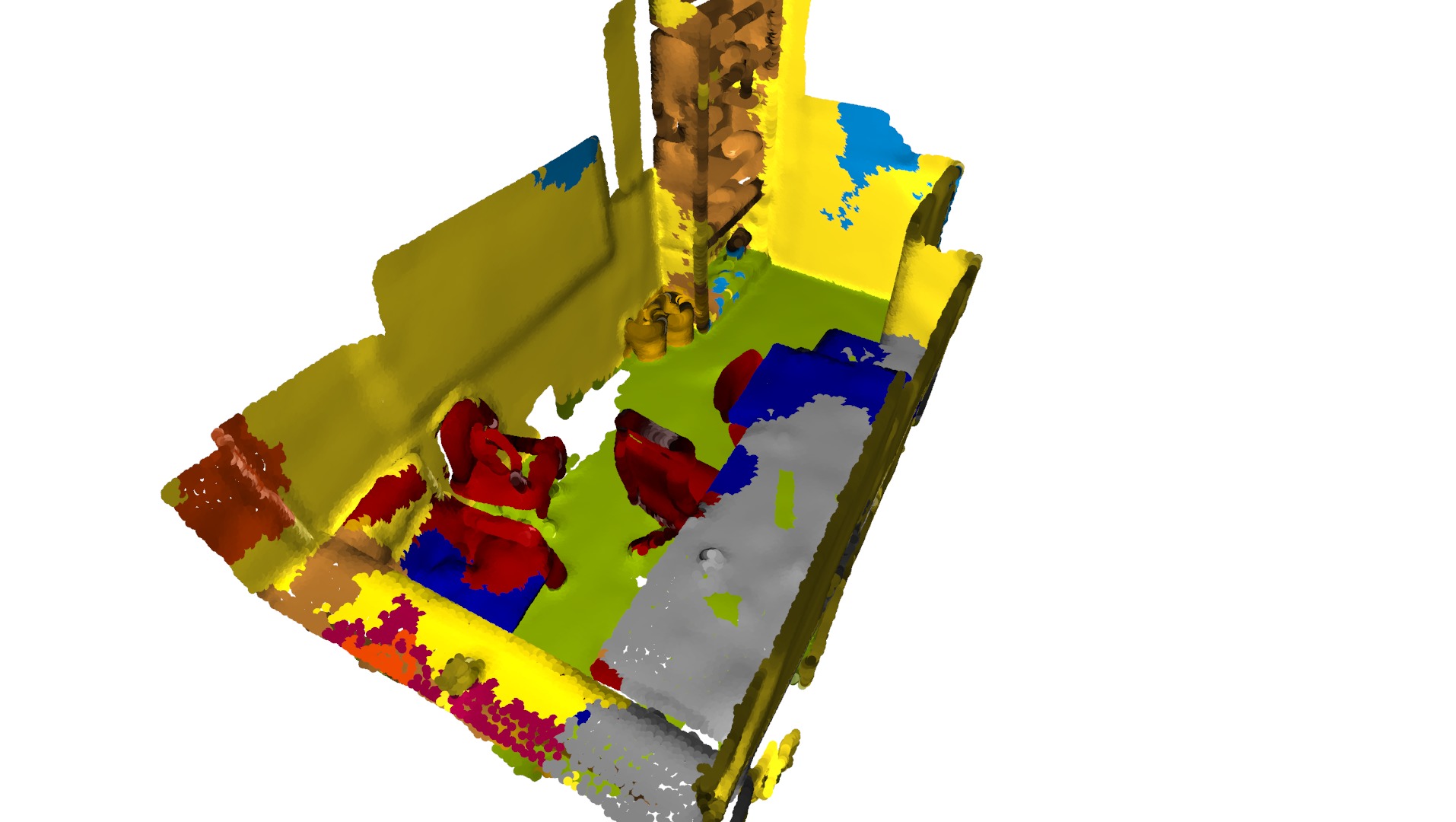}
    & \includegraphics[width={\wind}mm, totalheight={\hind}mm, trim={0 20mm 0 20mm}, clip]{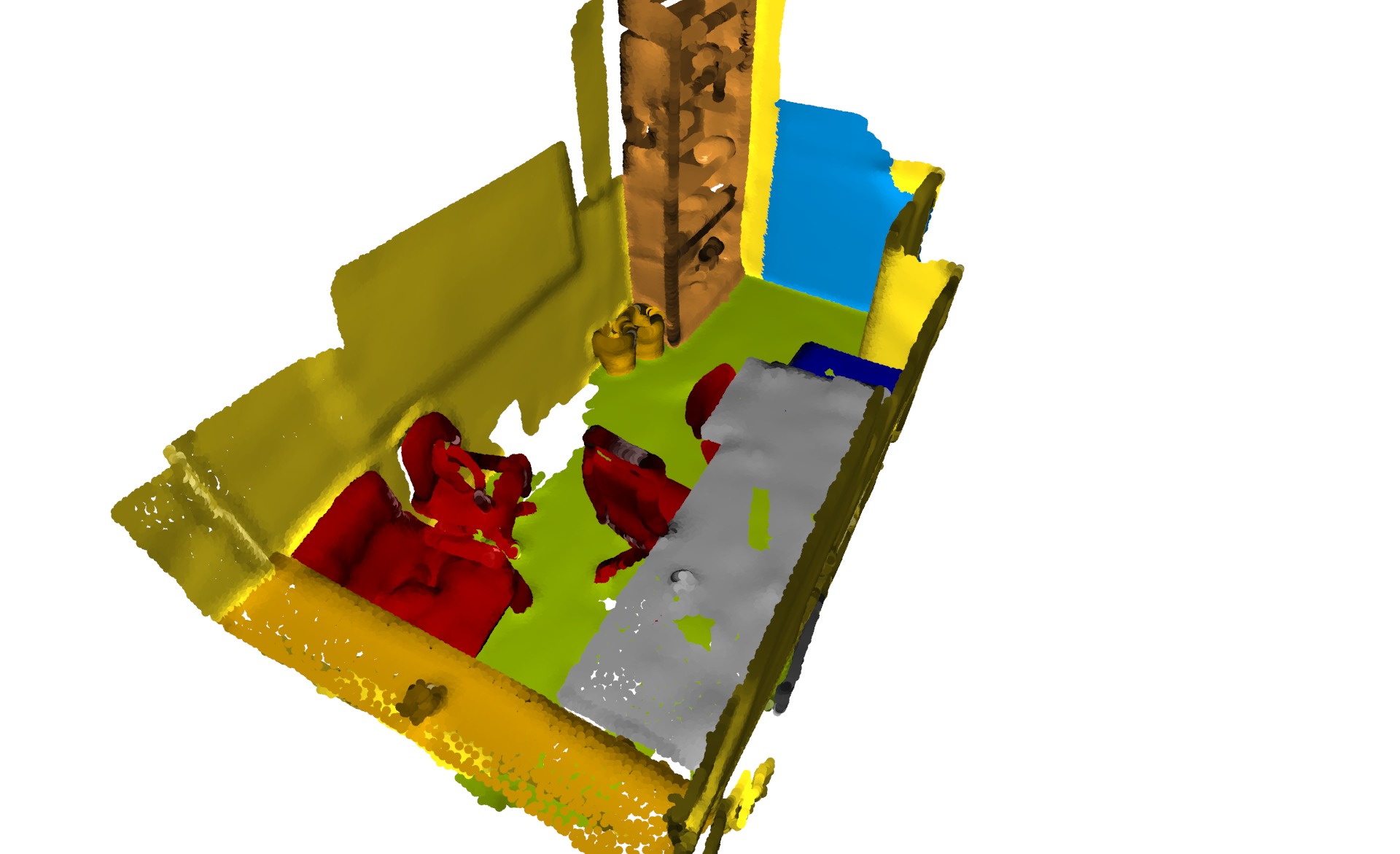} \\
    \includegraphics[width={\wind}mm, totalheight={\hind}mm, trim={0 20mm 0 20mm}, clip]{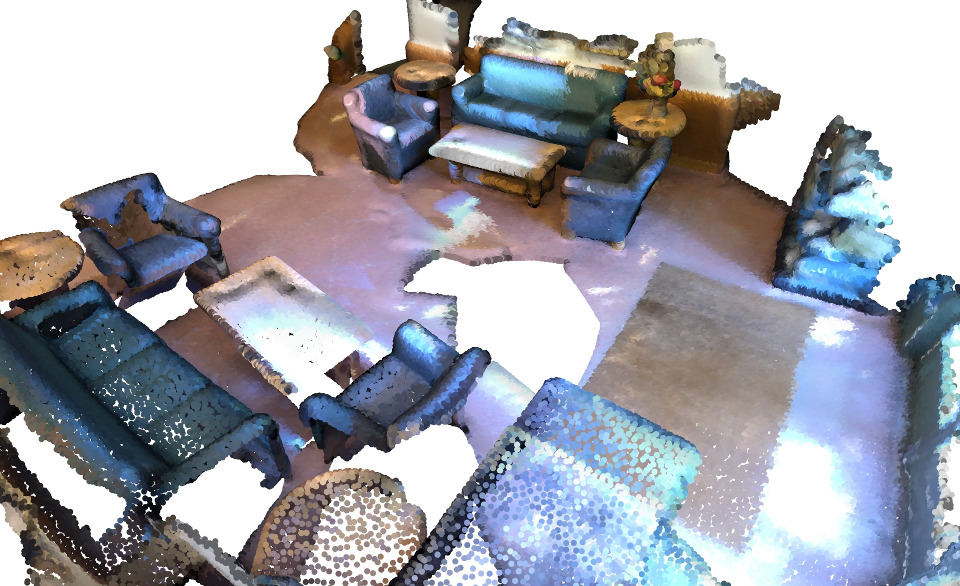}
    & \includegraphics[width={\wind}mm, totalheight={\hind}mm, trim={0 20mm 0 20mm}, clip]{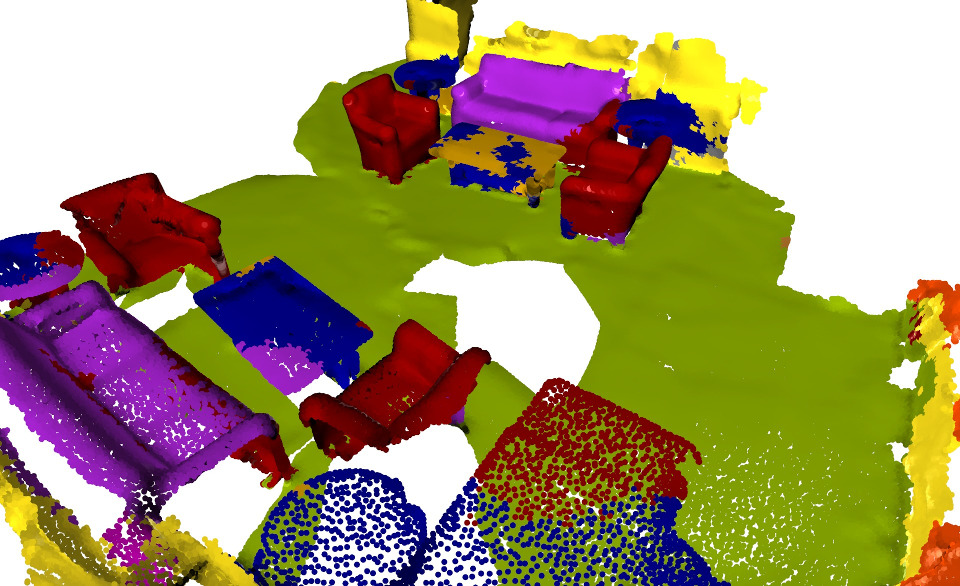}
    & \includegraphics[width={\wind}mm, totalheight={\hind}mm, trim={0 20mm 0 20mm}, clip]{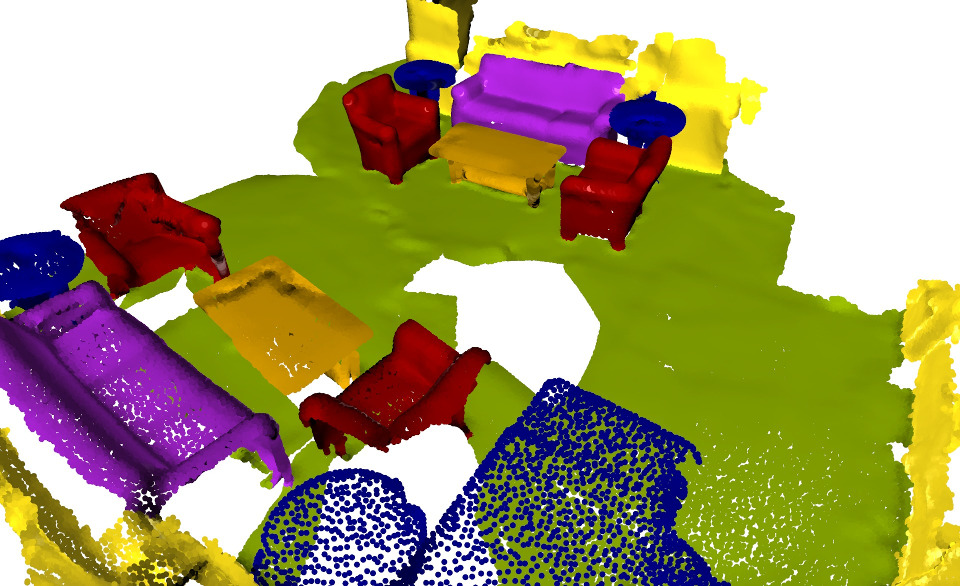} \\
	\multicolumn{3}{c}{
		\textcolor{scannet_1}{\ColorMapCircle} Wall
		\textcolor{scannet_2}{\ColorMapCircle} Floor
		\textcolor{scannet_3}{\ColorMapCircle} Cabinet
		\textcolor{scannet_4}{\ColorMapCircle} Bed
		\textcolor{scannet_5}{\ColorMapCircle} Chair
		\textcolor{scannet_6}{\ColorMapCircle} Sofa
		\textcolor{scannet_7}{\ColorMapCircle} Table
		\textcolor{scannet_8}{\ColorMapCircle} Door
		\textcolor{scannet_9}{\ColorMapCircle} Window
		\textcolor{scannet_10}{\ColorMapCircle} Bookshelf
		\textcolor{scannet_13}{\ColorMapCircle} Desk
		\textcolor{scannet_20}{\ColorMapCircle} Other furniture

	}\\\\
\end{tabular}
}
\caption{Qualitative results on ScanNet \cite{dai17}.}
\label{fig:results_scannet}
\end{figure*}

\begin{figure*}
\centering
\resizebox*{1\textwidth}{!}{
\begin{tabular}{@{}c@{\hspace{1mm}}c@{\hspace{1mm}}c@{\hspace{1mm}}}
	
	Color & Prediction & Ground truth\\
	\includegraphics[width={\wind}mm, totalheight={\hind}mm, trim={0 20mm 0 20mm}, clip]{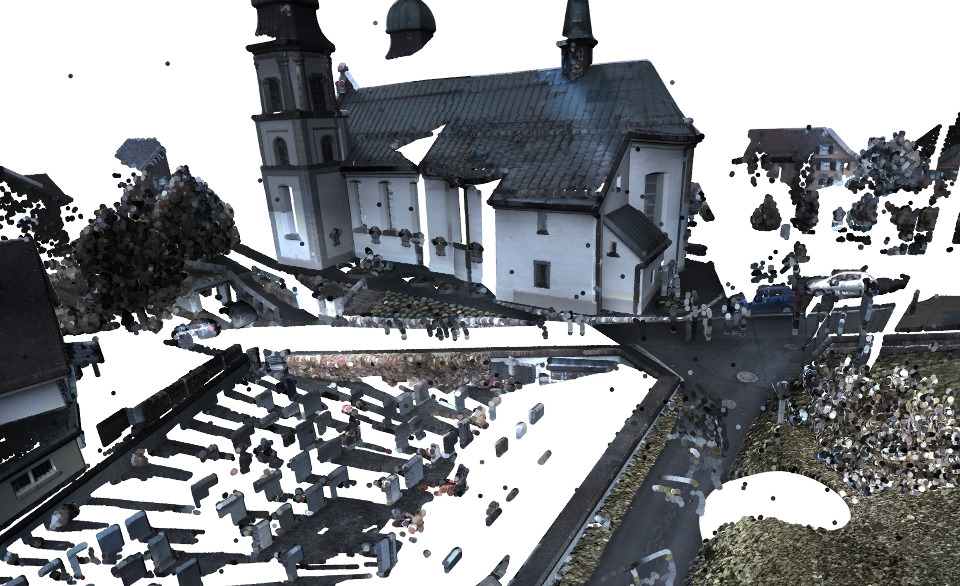}
    & \includegraphics[width={\wind}mm, totalheight={\hind}mm, trim={0 20mm 0 20mm}, clip]{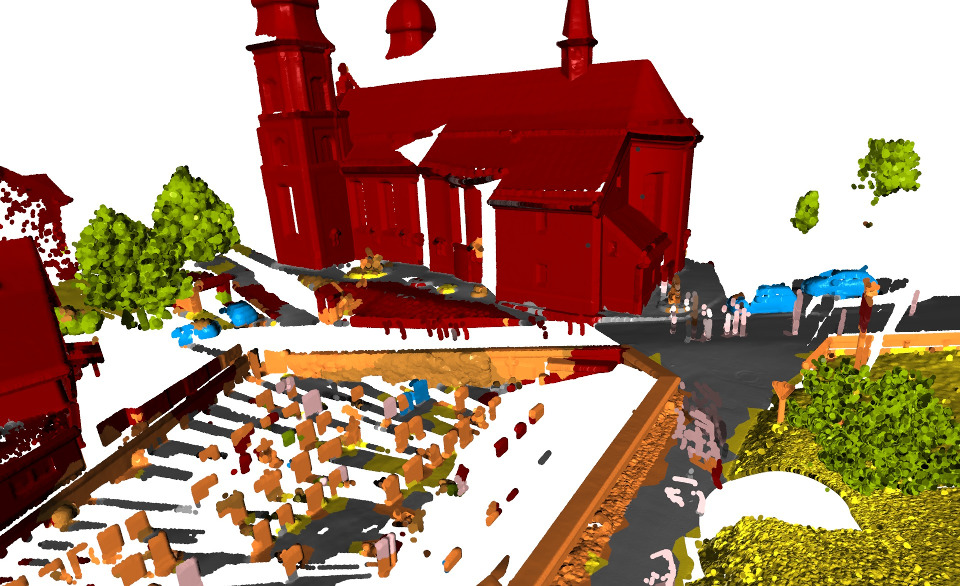}
    & \includegraphics[width={\wind}mm, totalheight={\hind}mm, trim={0 20mm 0 20mm}, clip]{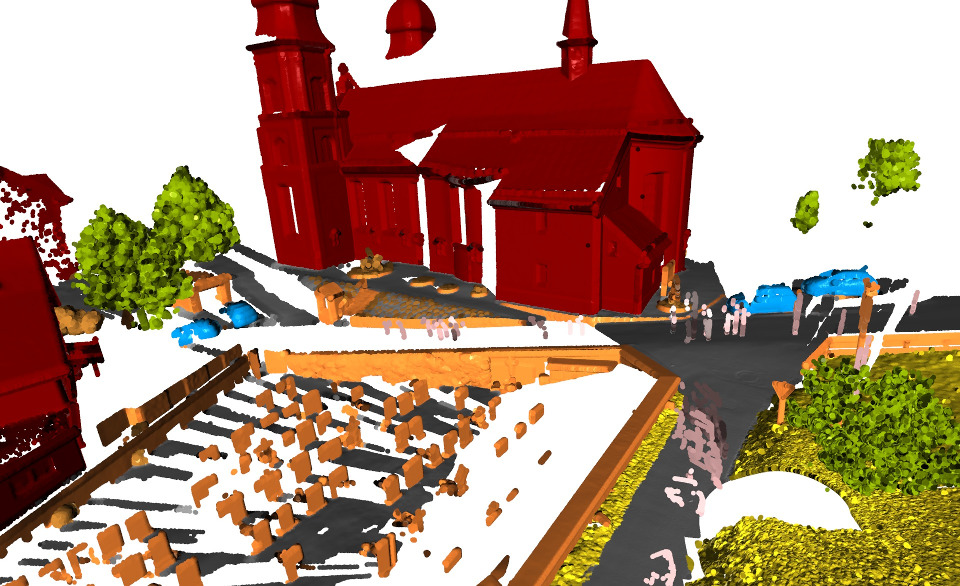} \\
    \includegraphics[width={\wind}mm, totalheight={\hind}mm, trim={0 20mm 0 20mm}, clip]{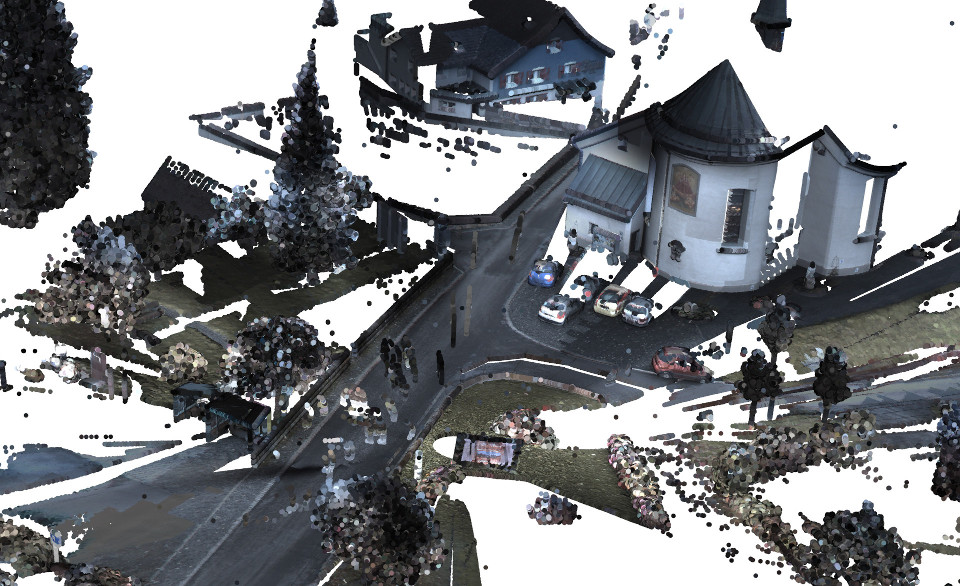}
    & \includegraphics[width={\wind}mm, totalheight={\hind}mm, trim={0 20mm 0 20mm}, clip]{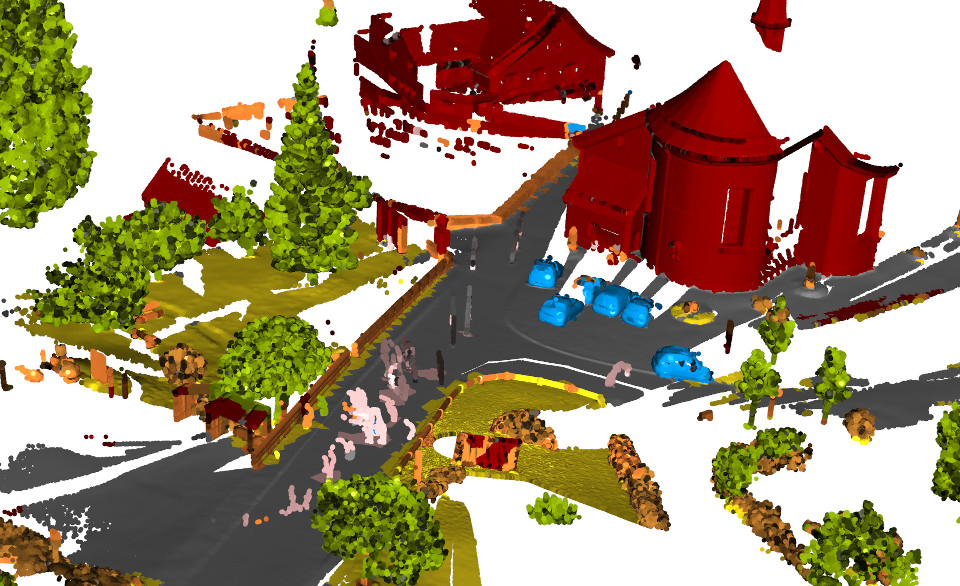}
    & \includegraphics[width={\wind}mm, totalheight={\hind}mm, trim={0 20mm 0 20mm}, clip]{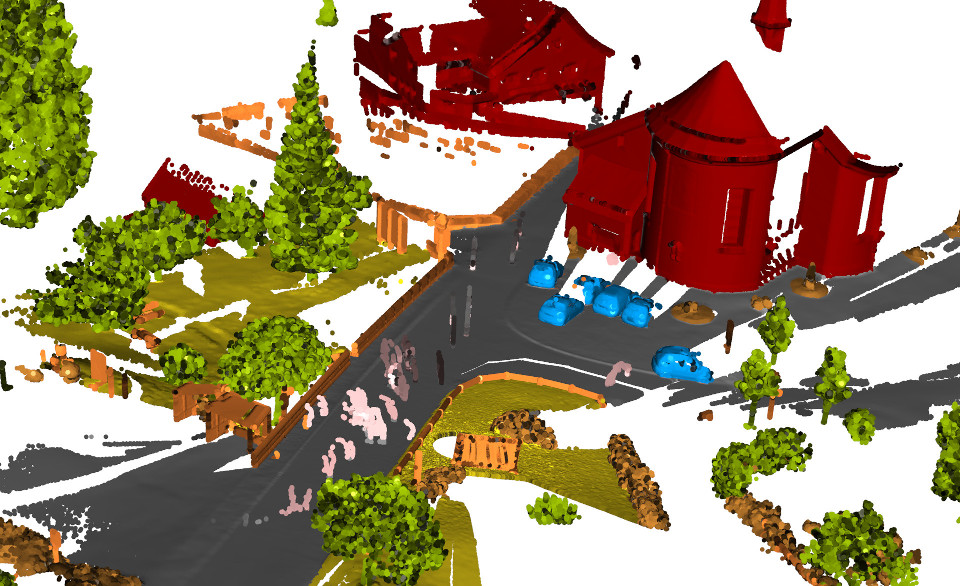} \\
    \includegraphics[width={\wind}mm, totalheight={\hind}mm, trim={0 20mm 0 20mm}, clip]{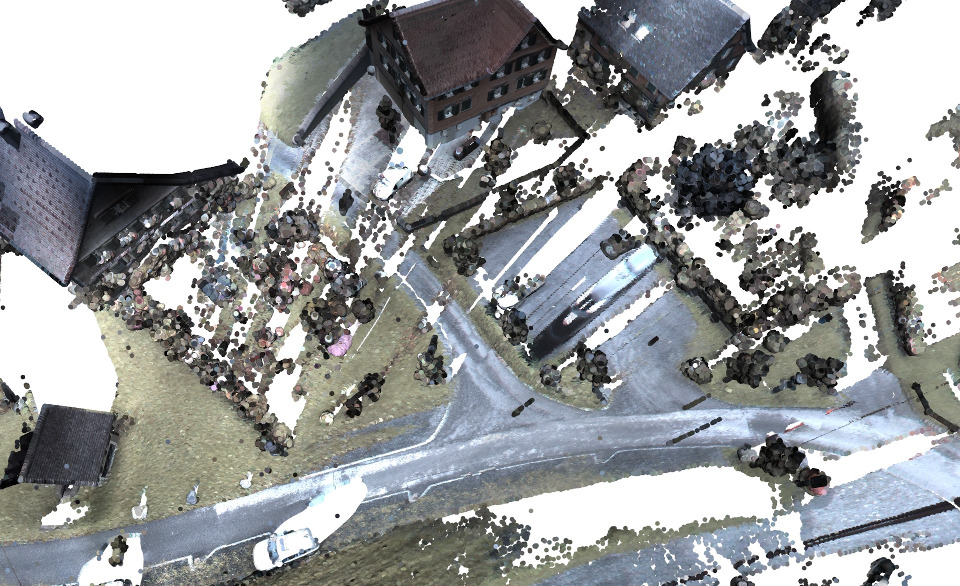}
    & \includegraphics[width={\wind}mm, totalheight={\hind}mm, trim={0 20mm 0 20mm}, clip]{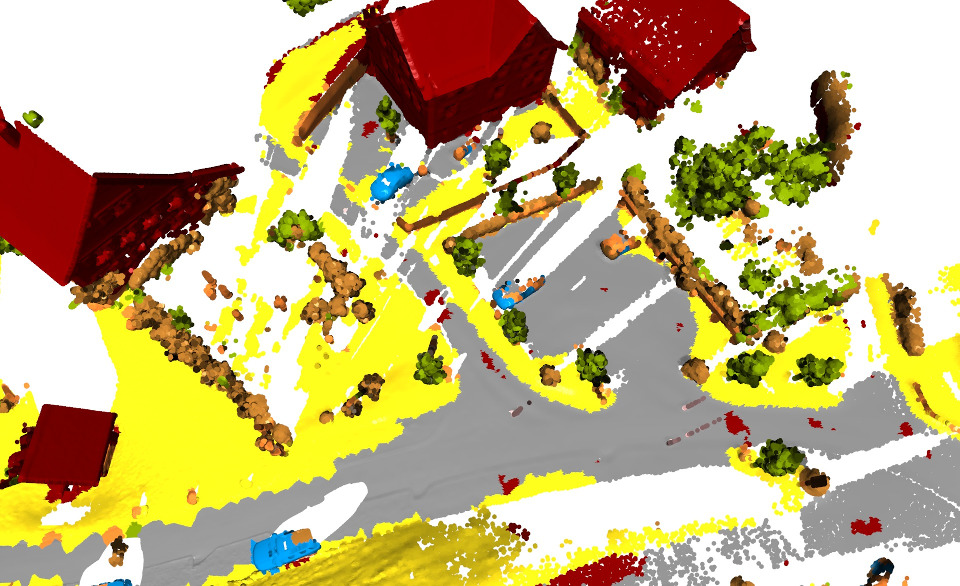}
    & \includegraphics[width={\wind}mm, totalheight={\hind}mm, trim={0 20mm 0 20mm}, clip]{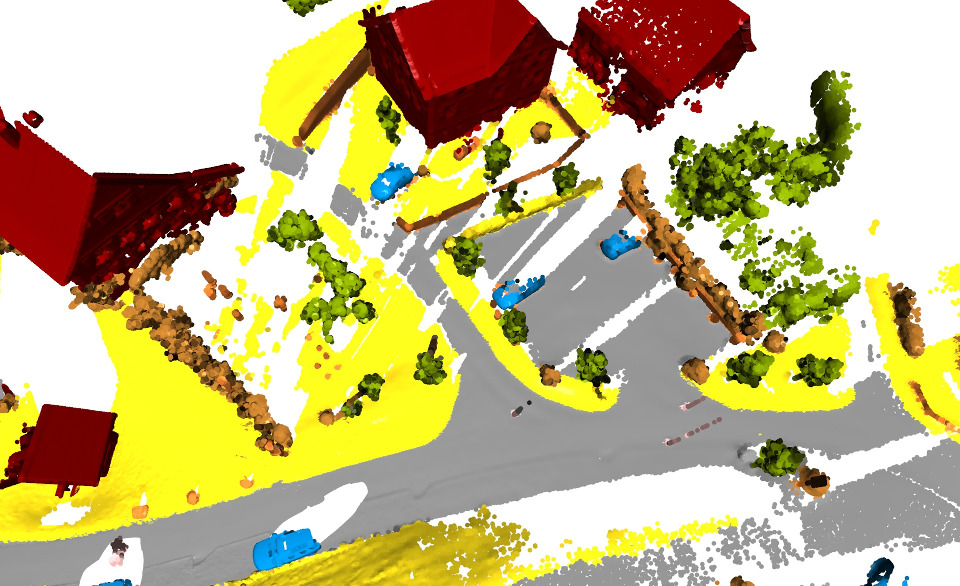} \\
	\multicolumn{3}{c}{
		\textcolor{sem8_1}{\ColorMapCircle} Man made terrain
		\textcolor{sem8_2}{\ColorMapCircle} Natural terrain
		\textcolor{sem8_3}{\ColorMapCircle} High vegetation
		\textcolor{sem8_4}{\ColorMapCircle} Low vegetation
		\textcolor{sem8_5}{\ColorMapCircle} Building
		\textcolor{sem8_6}{\ColorMapCircle} Hardscape
		\textcolor{sem8_7}{\ColorMapCircle} Scanning artifacts
		\textcolor{sem8_8}{\ColorMapCircle} Cars
	}\\\\
\end{tabular}
}
\caption{Qualitative results on Semantic3D \cite{hackel17}.}
\label{fig:results_semantic3d}
\end{figure*}

We provide more qualitative results of our method on different datasets in Figures \ref{fig:results_s3dis}-\ref{fig:results_semantic3d}.

\end{document}